\documentclass[letterpaper, 10 pt, conference]{ieeeconf}  %

\IEEEoverridecommandlockouts                              %

\usepackage[pagebackref=true]{hyperref}
\hypersetup{
  colorlinks=true,
  citecolor=green,
  linkcolor=red,
  urlcolor=magenta,
}

\usepackage{adjustbox}
\usepackage{amsfonts}
\usepackage{amsmath}
\usepackage{amssymb}
\usepackage{authblk}
\usepackage{booktabs}
\usepackage{capt-of}
\usepackage[font=small]{caption}
\usepackage{colortbl}
\usepackage{wrapfig}
\usepackage{comment}
\usepackage{epsfig}
\usepackage{etoolbox}
\usepackage{float}
\usepackage[T1]{fontenc}
\usepackage{gensymb}
\usepackage{lipsum}
\usepackage[ruled]{algorithm2e}
\usepackage{graphicx}
\usepackage[utf8]{inputenc}
\usepackage{listings}
\usepackage{mathtools}
\usepackage{microtype}
\usepackage{lipsum,booktabs}
\usepackage{multirow}
\usepackage{nicefrac}
\usepackage{paralist}
\usepackage{siunitx}  %
\usepackage{subcaption}
\usepackage{svg}
\usepackage{times}
\usepackage{tabularx}
\usepackage{textcomp}
\usepackage{titletoc}
\usepackage{wrapfig}
\usepackage{xcolor}
\usepackage{listings}

\definecolor{codegreen}{rgb}{0,0.6,0}
\definecolor{codegray}{rgb}{0.5,0.5,0.5}
\definecolor{codepurple}{rgb}{0.58,0,0.82}
\definecolor{backcolour}{rgb}{0.95,0.95,0.92}

\lstdefinestyle{mystyle}{
    backgroundcolor=\color{backcolour},   
    commentstyle=\color{codegreen},
    keywordstyle=\color{magenta},
    numberstyle=\normalsize\color{codegray},
    stringstyle=\color{codepurple},
    basicstyle=\ttfamily\normalsize,
    breakatwhitespace=false,         
    breaklines=true,                 
    captionpos=b,                    
    keepspaces=true,                 
    numbers=left,                    
    numbersep=5pt,                  
    showspaces=false,                
    showstringspaces=false,
    showtabs=false,                  
    tabsize=4
}

\newcommand{\mc}[1]{\mathcal{#1}}

\newcommand{\cf}{\emph{cf.}}

\renewcommand{\baselinestretch}{0.97}

\newtheorem{definition}{Definition}

\allowdisplaybreaks[1]
\renewcommand{\baselinestretch}{1.05}
\definecolor{flodarkpurple}{rgb}{0.288,0.1196,0.7}

\newcommand{\authorhref}[3][flodarkpurple]{\href{#2}{\color{#1}{#3}}}%

\newcommand{\coolname}{$f$-Cal}
\newtheorem{theorem}{Theorem}

\makeatletter
\renewcommand\AB@affilsepx{, \protect\Affilfont}
\makeatother

\title{\LARGE \bf
$f$-\textit{Cal}: Calibrated aleatoric uncertainty estimation from neural networks for robot perception %
}

\author[1]{\authorhref{http://dhaivat1729.github.io/}{Dhaivat Bhatt$^*$}\thanks{$^*$Equal contribution. Project page: 
\href{https://f-cal.github.io}{https://f-cal.github.io}}}
\author[1]{\authorhref{https://scholar.google.com/citations?user=MnPjDIgAAAAJ\&hl=en}{Kaustubh Mani}$^*$}
\author[1]{\authorhref{https://mila.quebec/en/person/dishank-bansal/}{Dishank Bansal}}
\author[1]{\authorhref{http://krrish94.github.io/}{Krishna Murthy}}
\author[2]{\authorhref{https://www.linkedin.com/in/lee-hanju-1848323/?originalSubdomain=jp}{Hanju Lee}}
\author[1]{\authorhref{http://liampaull.ca}{Liam Paull}}
\vspace{-0.5cm}
\affil[1]{\href{https://montrealrobotics.ca}{Montreal Robotics and Embodied AI Lab}, \href{https://mila.quebec/en}{Mila}, \href{https://diro.umontreal.ca/accueil/}{Universit\'e de Montr\'eal}}
\affil[2]{\href{https://www.denso.com/global/en/}{Denso}}

\begin{document}

\maketitle
\thispagestyle{plain}
\pagestyle{plain}

\begin{abstract}
While modern deep neural networks are performant perception modules, performance (accuracy) alone is insufficient, particularly for safety-critical robotic applications such as self-driving vehicles.
Robot autonomy stacks also require these otherwise blackbox models to produce reliable and \emph{calibrated} measures of confidence on their predictions.
Existing approaches estimate uncertainty from these neural network perception stacks by modifying network architectures, inference procedure, or loss functions.
However, in general, these methods lack \emph{calibration}, meaning that the predictive uncertainties do not faithfully represent the true underlying  uncertainties (process noise). %
Our key insight is that calibration is only achieved by imposing constraints across multiple examples, such as those in a mini-batch; as opposed to existing approaches which only impose constraints per-sample, often leading to overconfident (thus miscalibrated) uncertainty estimates.
By enforcing the distribution of outputs of a neural network to resemble a target distribution by minimizing an $f$-divergence, we obtain significantly better-calibrated models compared to prior approaches.
Our approach, \coolname{}, outperforms existing uncertainty calibration approaches on robot perception tasks such as object detection and monocular depth estimation over multiple real-world benchmarks.
\end{abstract}

\section{Introduction}
\label{sec:intro}

The \textit{performance} of deep neural network-based visual perception systems has increased dramatically in recent years. However, for safety-critical \textit{embodied} applications, such as autonomous driving, performance alone is not sufficient.
The absence of reliable and \emph{calibrated} uncertainty estimates in neural network predictions precludes us from incorporating these into downstream sensor fusion \cite{shin2018direct} or probabilistic planning \cite{romandhaivat, gopalakrishnan2017prvo, blackmore2006probabilistic} components.

The tools of probabilistic robotics require calibrated confidence/uncertainty measures, in the form of a \emph{measurement model} $z=h(x, \nu)$. For a traditional sensor, this model $h$ is specified by the designer's understanding of the physical sensing processes, and the noise distribution parameters $\nu$ are estimated by controlled calibration experiments with known ground truth states $x^*$ and sensor observations $z$.
However, for deep neural networks (DNNs) to be used as \emph{sensors} in typical robotic perception stacks, estimating the noise distribution is a much more challenging task for several reasons. First, the domain of inputs is extremely high dimensional (e.g., RGB images) - generating a calibration setup for every possible input is infeasible. Second, the noise distribution is input dependent (heteroscedastic). Finally, neural networks typically transform the inputs via millions of nonlinear operations, preventing approximation by simpler (e.g., piecewise affine) models.
We envision a \textbf{deep neural network (DNN) as a sensor} paradigm where a DNN outputs calibrated predictive distributions that may directly be used in probabilistic planning or sensor fusion. The challenge, however, is that these predictive distributions must be learned solely from training data, with neither additional postprocessing nor architectural modifications.

\begin{figure}[!t]
    \centering
    \includegraphics[width=\linewidth]{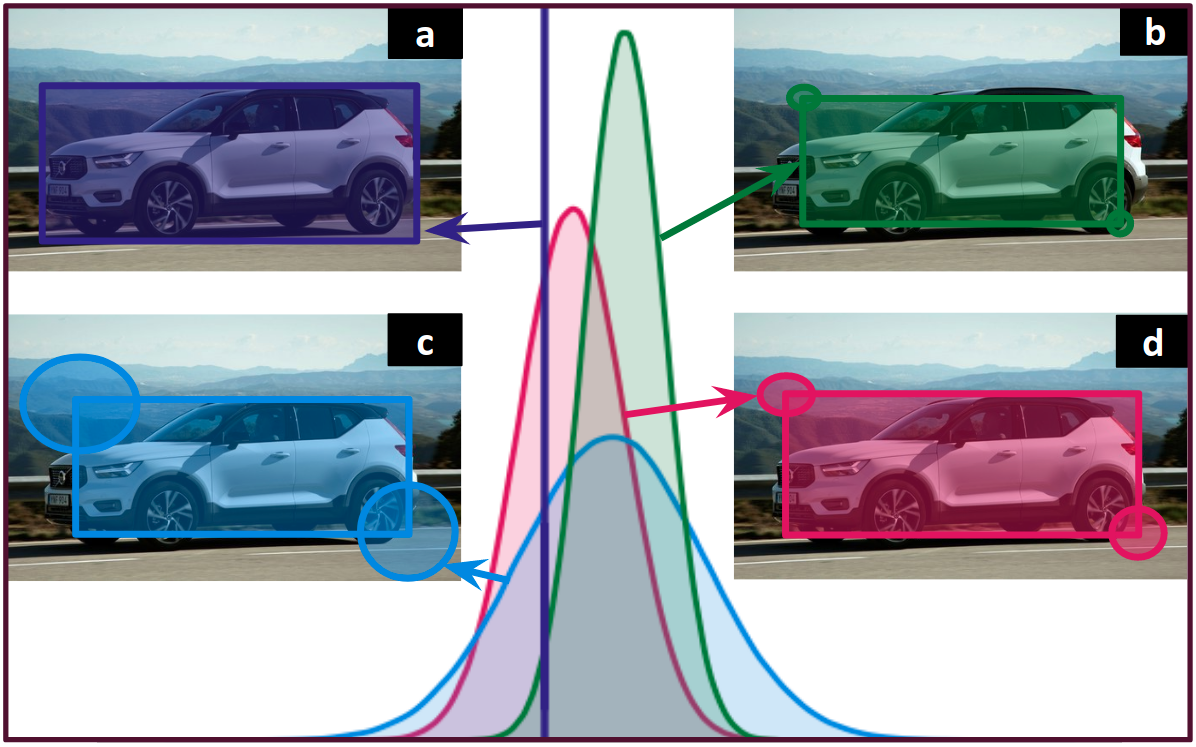}
    \caption{\textbf{\coolname{}} enables us to obtain \emph{calibrated} measures of uncertainty from otherwise blackbox neural networks used in robot perception tasks. This didactic example demonstrates how \coolname{} can estimate the \emph{aleatoric} uncertainty from object detectors. (a) depicts a ground-truth bounding box, a single-sample (dirac-delta) distribution; (b) and (c) denote \emph{uncalibrated} probabilistic outputs from a Bayesian neural network -- (b) is overconfident and inconsistent, (c) is consistent but underconfident; (d) denotes a calibrated estimate, i.e., the error ellipses correspond to the \emph{true} underlying aleatoric uncertainty.
    }
    \label{fig:intro_figure}
\end{figure}

Our key insight is that \textbf{distributional calibration cannot be achieved by a loss function that operates over individual samples}. This motivates a new loss function that enforces calibration through a distributional constraint that is imposed upon uncertainty estimates across multiple (i.i.d.) samples.
Specifically, our approach \coolname{}, minimizes an $f$-divergence between a specified canonical distribution and an empirical distribution generated from neural network predictions, as shown in Fig. \ref{fig:pipeline}.
Unlike prior approaches~\cite{gp-beta, isotonic, bosch-calib}, we neither require a held-out calibration dataset nor impose any inference time overhead.
For a given performance threshold, \coolname{} achieves better calibration compared to current art.
We demonstrate the effectiveness, scalability and widespread applicability of this approach on large-scale, real-world tasks such as object detection and depth estimation.

\vspace{-0.2cm}
\section{Related Work}
\label{sec:relatedwork}
\vspace{-0.2cm}
The rapidly growing field of \textbf{Bayesian deep learning} has resulted in the development of models that estimate a \emph{distribution} over the output space~\cite{gal2016uncertainty,gal2016dropout, kendall2017uncertainties,lakshminarayanan2017simple, blundell2015weight}. There is a distinction between uncertainty that is due to the stochasticity of the underlying process (\emph{aleatoric}) versus uncertainty that is due to the model being insufficiently trained (\emph{epistemic})~\cite{kendall2017uncertainties}. 
\textbf{Epistemic} uncertainty is often estimated by either using ensembles of neural networks or by stochastic regularization at inference time (Monte-Carlo dropout)~\cite{gal2016dropout, lakshminarayanan2017simple, tagasovska2019single}.
\textbf{Distributional} uncertainty is also being extensively studied, to detect out of training-distribution examples~\cite{hendrycks2016baseline, lakshminarayanan2017simple, hein2019relu,liang2017enhancing, guo2017calibration, hendrycks2018deep, mohseni2020self, malinin2018predictive,  sehwag2019analyzing}. However, there is no direct approach to address distributional uncertainty for regression settings.

In this work, we assume distributional and epistemic uncertainty to be low (i.e., in-distribution setting with reasonably well-trained models such as those common in robot perception), and focus specifically on calibrating \textbf{aleatoric uncertainty} estimates in regression problems. Such challenging settings have received far less attention in terms of uncertainty estimation~\cite{kuleshov2018accurate, gp-beta, levi2019evaluating, bosch-calib}. Existing calibration techniques are post-hoc and either require a large held-out calibration dataset~\cite{bosch-calib} and/or add parameters to the model after training~\cite{bosch-calib, isotonic}.
Quantile regression methods~\cite{chung2020beyond,tagasovska2019single, ho2005calibrated, rueda2007calibration, taillardat2016calibrated} quantify uncertainty by the fraction of predictions in each quantile.
Other methods, such as isotonic regression and temperature scaling, have also been extended to be the regression setting \cite{kuleshov2018accurate, bosch-calib}. Authors in \cite{gast2018lightweight} proposed an alternate architecture for aleatoric uncertainty estimation. However, \coolname{} is completely architecture agnostic, and can be applied to any probabilistic neural regressors. More recently, a \emph{calibration loss} is proposed in \cite{bosch-calib} that enforces the predicted variances to be equal to per-sample errors, thus grounding each prediction. However, this takes on a \emph{local view} of the calibration problem, and while individual samples might appear well-calibrated, the overall distribution of the regressor errors exhibits a strong deviation from the expected target distribution (\cf Sec.~\ref{sec:results}).

A recent approach that is somewhat similar to ours in spirit is Gaussian process beta calibration (GP-beta)~\cite{gp-beta}. It is a post-hoc approach that employs a Gaussian process model (with a beta-link function prior) to calibrate uncertainties during inference. This requires the computation of pairwise statistics, exacerbating inference time. In the maximum mean discrepancy (MMD) loss \cite{cui2020calibrated} distribution matching is performed to achieve calibration. This method was proposed for small datasets and does not scale well with input size. \coolname{} is a superior performing loss function that requires the same inference time as typical Bayesian neural networks~\cite{kendall2015bayesian}.

\section{Problem Setup}
\label{sec:background}
\subsection{Preliminaries}
\label{sec:preliminaries}
We assume a regression problem over an i.i.d. labelled training dataset 
$\mathcal{D} \triangleq \{(x_i,y_i)\}_{i = 1 \ldots |\mc{D}|}$ with $x_i \in \mathcal{X}$ where $\mathcal{X}$ is the ($n$-dimensional) input space and $y_i \in \mathcal{Y}$ where $\mc{Y} \subseteq \mathbb{R}^n$ is the output space.
A \textit{deterministic} model $f_{d}:\mathcal{X}\mapsto\mathcal{Y}$~\footnote{In practice these models are assumed to be neural networks with parameters $\theta$ but we omit the $\theta$ for clarity at this stage.} directly learns the mapping from the input to the output space by minimizing a loss function $\mathcal{L}: \mathcal{Y}\times\mathcal{Y} \mapsto \mathbb{R}$, for example through empirical risk minimization:
\begin{equation}
    R_{emp}(f_d) = \frac{1}{N}\sum_{i=1}^N \mathcal{L}(f_d(x_i),y_i).
    \label{eq:empirical_risk}
\end{equation}
Equation \ref{eq:empirical_risk} is typically estimated over a mini-batch of size $N<<|\mc{D}|$  during stochastic gradient descent (SGD). Following the notation in \cite{gp-beta}, we desire a \textit{probabilistic} model $f_p:\mathcal{X}\mapsto \mc{S_Y}$ where $\mc{S_Y}$ is the space of all probability density functions $s(y)$ over $\mc{Y}$ ($s: \mc{Y} \mapsto [0,\infty)$ and $\int s(y)dy = 1$). The probability density function (PDF) is defined through its cumulative density function (CDF): $S(y) = \int_{-\infty}^y s(y')dy'$. 
\subsection{Uncertainty Calibration}

Calibrated uncertainty estimates are those where the output uncertainties can be exactly interpreted as confidence intervals of the underlying target label distribution.
This allows uncertainty estimates across multiple samples (and models) to be compared. Intuitively, we understand the notion of uncertainty calibration to mean that if we repeated a stochastic experiment a large number of times, for example by asking many different people to label the same image, that the ``label generating distribution'' matches the predictive distribution of the model:
\vspace{-0.3cm}
\begin{equation}
    y_i \sim f_p(x_i)
    \label{eq:y}
\end{equation}

However, it is impractical to label every piece of data multiple times. Instead, we aggregate the labels across many different inputs to produce calibrated predictive distributions.  Using our definitions from Sec. \ref{sec:preliminaries} and adapting from \cite{gp-beta}, we can define what we desire in terms of calibration in the case of a deep neural regressor as follows:

\begin{figure*}[!ht]
    \centering
    \includegraphics[width=\textwidth]{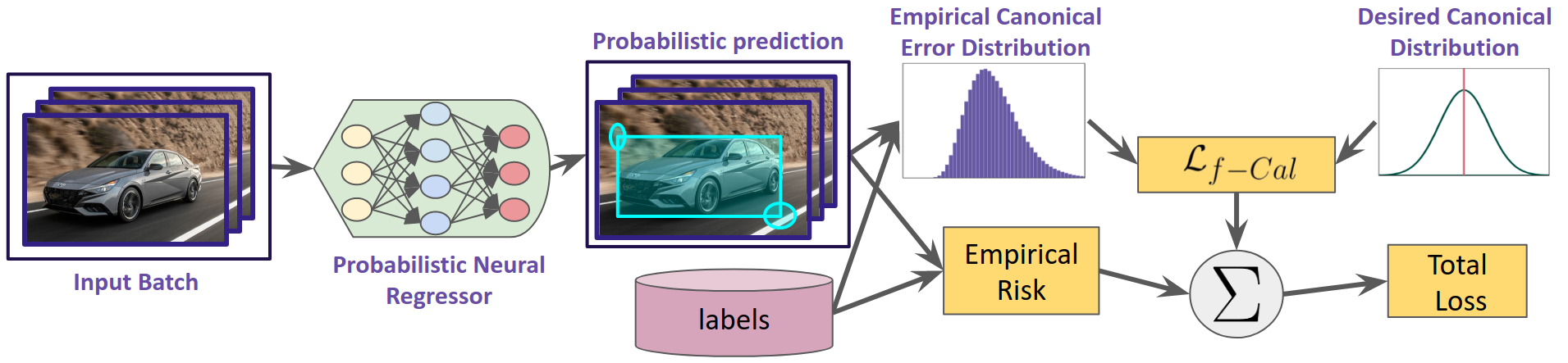}
    \caption{\textbf{\coolname{} pipeline}: We make a conceptually simple tweak to the loss function in a typical (deterministic) neural network training pipeline. In addition to the empirical risk (e.g., $L1$, $L2$, etc.) terms, we impose a distribution matching constraint ($\mathcal{L}_{f-\textnormal{Cal}}$) over the error residuals across a mini-batch.
    By encouraging the distribution of these error residuals to match a target \emph{calibrating distribution} (e.g., Gaussian), we ensure the neural network predictions are \emph{calibrated}.
    Compared to prior approaches, most of which perform post-hoc calibration, or require large held-out calibration datasets, \coolname{} does not impose an inference time overhead.
    \coolname{} is task and architecture agnostic, and we apply it to robot perception problems such as object detection and depth estimation.
    }
    \label{fig:pipeline}
\vspace{-0.3cm}
\end{figure*}

\begin{definition}[\textbf{Uncertainty Calibration}]
A neural regressor $f_p$ is calibrated if and only if \footnote{Referring to Fig.~\ref{fig:intro_figure}, the requirement for calibration is more stringent than that of consistency, which is a one-way constraint at an arbitrary confidence bound $c$:  $p(Y \leq y | s(y)) \leq c$ }:
\begin{equation}
    p(Y \leq y | s(y)) = \int_{-\infty}^y s(y')dy' \text{  } \forall y \in \mc{Y}
\end{equation}
\label{definition:calibration}
\end{definition}
In the above definition, $Y$ is an instantiation of the random variable $y$. If we can assume that the noise is sampled from a parametric distribution $s(y;\phi)$, then the probabilistic model need only output the parameters associated with each sample. 
In this case, we can consider the model to be calibrated if and only if the aggregated error statistics over multiple outputs of a model align with the parameters predicted by the model.

\subsection{Loss Attenuation (Negative Log-Likelihood - NLL)}

The most widely used technique for estimating heteroscedastic aleatoric uncertainty is \textit{loss attenuation} \cite{kendall2017uncertainties, loss_att}, which performs maximum likelihood estimation by minimizing the negative log-likelihood loss:
\begin{align}
    R_{emp}(f_p) & = - \frac{1}{N}\sum_{i=1}^{N} \mathcal{L_{\text{LA}}}(f_p(x_i),y_i) \nonumber \\ 
    & = - \frac{1}{N}\sum_{i=1}^{N} \log{s(y_i; f_p(x_i))}
\end{align}
\noindent
since $f_p(x_i)$ outputs the parameters of the distribution.
For example, if the aleatoric uncertainty is characterized by a Gaussian random variable ($\phi \triangleq (\mu,\sigma)$), the above expression becomes
\begin{equation}
    R_{emp}(f_p) = \frac{1}{N} \sum_{i = 1}^{N} \frac{1}{2} \left( \frac{(y_i - \mu_i)^2}{\sigma_i^2}  + \log{\sigma_i^2} \right)
    \label{equation:nll}
\end{equation}
We refer to the loss in \eqref{equation:nll} as the \textbf{NLL} loss in the experiments. However, probabilistic neural regressors trained using this NLL objective typically lack \textbf{calibration} according to Def.~\ref{definition:calibration}. %

\section{\coolname{}: Variational Inference for Aleatoric Uncertainty Calibration}
\label{sec:approach}
In this section we present \coolname{}, a principled approach to obtain calibrated aleatoric uncertainty from neural nets.

\subsection{Calibration as Distribution Matching}

Following the definition of distributional calibration (Def.~\ref{definition:calibration}), \coolname{} formulates a variational minimization objective to calibrate the uncertainty estimates from a deep network.

In the case of a traditional (non deep learning based) sensor, we would calibrate the noise distribution with the procedure:

\begin{enumerate}
    \item Choose a distribution family for the noise 
    \item For a fixed and known input value, draw multiple samples of the output observations 
    \item Fit the output samples to the distribution family    
\end{enumerate}

In the DNNS case, we only have one sample for any given input and we have no knowledge of the ground truth (noise free) label. We can similarly choose a distribution family for our model, but we cannot assume that any of the parameters are fixed across samples. Our approach to overcome this problem will be to assume that there is some canonical element of the distribution family that we can transform each predictive distribution to. Specifically, we seek to approximate the empirical posterior over some canonical transformation of the target variables $Y$ by a simpler (tractable) target distribution $Q$ (modeling choice).
This enables us to leverage an abundant class of distribution matching metrics, $f$-divergences, to formulate a loss function enforcing distributional calibration.
For tractable inference, we assume i.i.d. mini-batches of training data and instead impose distribution matching losses over empirical error residuals across each batch%
.

We assume that we can transform each training sample output distribution to some canonical element of the distribution family. For instance, Gaussian random variables are canonicalized by centering the distribution (subtracting the output label), followed by normalization (scaling the result by the inverse variance). These canonical elements are used (in conjunction with the labels) to determine the \emph{empirical} error distribution.
\coolname{} then performs distribution matching across this empirical and a target distribution.

\subsection{\coolname{} Algorithm}

Given a mini-batch containing $N$ inputs $x_i$, a probabilistic regressor predicts $N$ sets of distributional parameters $f_p(x_i) = \phi_i$ ($\phi_i \in \Phi)$ to the corresponding probability distribution $s(y_i; \phi_i)$. Define $g: \mc{Y} \times \Phi \mapsto \mc{Z}$ as the function that maps the target random variable $y_i$ to a random variable $z_i$ which follows a known canonical distribution. Since these residuals $\{z_1, z_2, \ldots,  z_N\}$ must ideally follow a chosen calibrating (target) distribution $Q$:
\begin{equation}
    z_i = g(y_i, \phi_i) \sim Q
    \label{eq:z}
\end{equation}
The key difference between \eqref{eq:y} and \eqref{eq:z} is that \eqref{eq:z} now applies \textbf{for all samples} in the dataset, as opposed to just a single sample. As a result, we can now follow the similar procedure that we would with a traditional sensor and compute the empirical statistics of the residuals of the $z_i$ variables across the entire set (or in practice across a mini-batch) to fit a proposal distribution $P_z$, and minimize the distributional distance from the canonical distribution $Q$.
This minimization can be performed with variational loss function that minimizes an $f$-divergence, $D_f(P_z||Q)$, between these two distributions. 
In summary, we propose a distribution matching loss function that augments typical supervised regression losses, and results in the neural regressor being calibrated to the target distribution:

\begin{align}
\label{eq:fcal}
    \mathcal{L} &=  (1 - \lambda) R_{emp}(f_p) + \lambda \mathcal{L}_{\text{\coolname{}}}  \\
    & = (1 - \lambda) R_{emp}(f_p)  +  \lambda D_f(P_z || Q) \nonumber
\end{align}

\noindent
where $\lambda$ is a hyper-parameter to balance the two loss terms (we provide thorough analysis of the choice of $\lambda$ in Sec.~\ref{sec:results}.
We experiment with a number of $f$-divergence choices, and identify KL-divergence and Wasserstein distance as viable choices.
Importantly, \coolname{} is agnostic to the choice of probabilistic deep neural regression model or task. In practice, it is a straightforward modification to the training loss function that can also be applied as a fine-tuning step to a previously partially trained model. 

\subsection{\coolname{} for Gaussian calibration}
\label{sec:gaussian-calibration}
The \coolname{} framework is generic and can be applied to arbitrary distributions. In this section we consider the case when the distribution $s(y_i;\phi_i)$ is Gaussian with $\phi_i \triangleq (\mu_i,\sigma_i)$. The variance $\sigma_i^2$ denotes the aleatoric uncertainty in this case.  The error residuals are computed as $z_i = \frac{y_i - \mu_i}{\sigma_i}$, where $\mu_i$ and $\sigma_i$ are predicted mean and the standard deviation of the $i$th Gaussian output from the neural network for each input $x_i$. So, $y_i \sim \mathcal{N}(\mu_i,\sigma_i^2)$, then $z_i \sim \mathcal{N}(0,1)$.

\begin{algorithm} [tb]
\SetCustomAlgoRuledWidth{0.6\textwidth}  %
\SetAlgoLined
\SetKwInOut{KInput}{Input}
\DontPrintSemicolon
\KInput{Dataset $D$, probabilistic neural regressor, $f_p$, degrees of freedom $K$, batch size $N$, number of samples for hyper-constraint $H$}
\For{$i = 1 \ldots N$}{
$(\mu_i, \sigma_i) \gets f_p(x_i)$ \\
$z_i \gets \frac{y_i - \mu_i}{\sigma_i}$ \\
}
$C = \emptyset$ \tcp*{\textnormal{Samples from Chi-squared distribution}}
\For{$i = 1 \ldots H$}{
    \tcp{\textnormal{Create a chi-squared hyper-constraint}}
    $\displaystyle q_i \gets \sum_{j = 1}^{K} z_{ij}^2,  z_{ij} \thicksim \{z_1 \cdot \cdot \cdot z_N\}$ \\
    C.append($q_i$)
}
$P_z \gets \textnormal{Fit-Chi-Squared-Distribution}(C)$ \\
$\mathcal{L}_{\text{\coolname{}}} \gets D_f(P_z || \chi_K^2)$ \\
\Return $\mathcal{L}_{\text{\coolname{}}}$
\caption{\coolname{} for Gaussian uncertainties}
\label{algorithm-gaussian}
\end{algorithm}
Optionally, one may apply several transforms to the random variables $y_i$ and impose distributional \emph{hyper-}constraints over the transformed variables. In practice, we find that this can improve the stability of the training process and enforces more stable calibration. In this case we compute the sum-of-squared error residuals $q = \sum_{i=1}^{K} z_i^2$, and enforce the resulting distribution to be Chi-squared with parameter $K$ i.e $q \sim \chi^2_K$, so in this case target distribution $Q = \chi_K^2$. Subsequently, we note that as the degrees of freedom $K$ of a Chi-squared distribution increase, it can be approximated by a Gaussian of mean $K$ and variance $2K$ through the application of the central limit theorem:
\begin{align*}
    &\lim_{K \to \infty} \frac{\chi^2_K -K}{\sqrt{2K}} \to \mathcal{N}(0,1) 
    \implies \lim_{K \to \infty} \chi^2_K \to \mathcal{N}(K,2K) 
\label{cs_central_limit_theorem}
\end{align*}

In practice, this variation of the central limit theorem for Chi-squared random variables holds for moderate values of $K$ (i.e., $K > 50$). This is practical to ensure, particularly in dense regression tasks such as bounding box object detection (where hundreds of proposals have to be scored per image) and per-pixel regression. We summarize the process for generating the calibration loss in Alg. \ref{algorithm-gaussian}. This loss is then combined with the typical empirical risk as given by \eqref{eq:fcal}.

\section{Experiments}
\label{sec:results}

We conduct experiments on a number of large-scale perception tasks, on both synthetic and real-world datasets. We report the following key findings which we elaborate on in the remainder of this section.
\begin{compactenum}
\item \coolname{} achieves significantly superior calibration compared to existing methods for calibrating aleatoric uncertainty.
\item These performance trends are consistently observed across multiple regression setups, neural network architectures, and dataset sizes.
\item We demonstrate that there is a trade-off between deterministic and calibration performance by varying the $\lambda$ hyper-parameter. This trade-off has been established in previous literature \cite{bosch-calib, guo2017calibration}. However, we further demonstrate empirically that this trade-off is inherently caused by a mismatch between the choice of the noise data distribution family and the true underlying noise distribution.  
\end{compactenum}

\subsection{Regression tasks}
We consider 3 regression tasks: a synthetic disc tracking dataset (Bokeh), KITTI depth estimation~\cite{kitti} and KITTI object detection~\cite{kitti}. These tasks are chosen to span the range of regression tasks relevant for robotics applications: sparse (one output per image in disc tracking), semi-dense (object detection), and pixelwise (fully) dense (depth estimation).
Unless otherwise specified, we model aleatoric uncertainty using heteroscedastic Gaussian distributions.

\subsection{Baselines}
We compare \coolname{} models with the following baselines: \textbf{NLL loss}~\cite{loss_att, gal2016uncertainty}, \textbf{Temperature scaling}~\cite{bosch-calib}, \textbf{Isotonic regression}~\cite{isotonic}, \textbf{Calibration loss}~\cite{bosch-calib} and \textbf{GP-beta}~\cite{gp-beta}. We report results for  \coolname{}, with KL-divergence (\textbf{\coolname{}-KL}) and Wasserstein distance (\textbf{\coolname{}-Wass}) as losses for distribution matching. 
We also experimented with a recently proposed maximum mean discrepancy based method~\cite{cui2020calibrated}. Being designed for very low data regimes, it failed to solve any of our tasks considered.
Similarly, GP-Beta~\cite{gp-beta} and isotonic regression \cite{isotonic} solve our synthetic tasks, but do not scale to large, real-world tasks.

\vspace{0.3cm}
\begin{table*}[ht]
\centering
\resizebox{\textwidth}{!}{%
\begin{tabular}{c|ccccc|ccccc|cccc|cccc}
                              & \multicolumn{5}{c|}{\textbf{Bokeh - synthetic dataset  (a)}}                                 & \multicolumn{5}{c|}{\textbf{KITTI - depth estimation (b)}}                        & \multicolumn{4}{c|}{\textbf{KITTI - Object detection (c)}}      & \multicolumn{4}{c}{\textbf{Cityscapes - Object detection (d)}} \\ \hline
\textbf{Approach}             & \textbf{L1(GT)$\downarrow$} & \textbf{L1$\downarrow$} & \textbf{ECE(z)$\downarrow$} & \textbf{ECE(q)$\downarrow$} & \textbf{NLL$\downarrow$} & \textbf{SiLog$\downarrow$} & \textbf{RMSE$\downarrow$} & \textbf{ECE(z)$\downarrow$} & \textbf{ECE(q)$\downarrow$} & \textbf{NLL$\downarrow$} & \textbf{mAP$\uparrow$} & \textbf{ECE(z)$\downarrow$} & \textbf{ECE(q)$\downarrow$} & \textbf{NLL$\downarrow$} & \textbf{mAP$\uparrow$} & \textbf{ECE(z)$\downarrow$} & \textbf{ECE(q)$\downarrow$} & \textbf{NLL$\downarrow$} \\ \hline
\textbf{NLL Loss\cite{loss_att}}              & 1.44                  & 1.54              & 1.73            & 91.83           & -1.60        & 9.213          & 2.850         & 2.39            & 99.0            & 2.403        & 54.451       & 0.304           & 5.37            & 1.022        & 38.309       & 0.224           & 3.503           & 1.069        \\ 
\textbf{Calibration  Loss\cite{bosch-calib}}    & 1.46                  & 1.57              & 1.13            & 76.11           & -1.68        & 9.604          & 2.918         & 1.71            & 99.9            & 2.879        & 50.405       & 2.33            & 81.848          & 0.773        & 39.218       & 0.163           & 9.681           & 0.999        \\ 
\textbf{Temperature  scaling\cite{bosch-calib}}  & 1.44                  & 1.54              & 0.82            & 9.22            & -1.70        & 9.213          & 2.850         & 2.36            & 99.9            & 3.362        & 54.451       & 0.315           & 4.151           & 1.021        & 38.309       & 0.226           & 2.705           & 1.065        \\ 
\textbf{Isotonic regression\cite{isotonic}}   & 1.38                  & 1.49              & 2.05            & 9.38            & -1.57        & -              & -             &                 & -               & -            & -            & -               & -               & -            & -            & -               & -               & -            \\ 
\textbf{GP-Beta\cite{gp-beta}}             & 1.39                  & 1.49              & 2.21            & 93.48           & -1.54        & -              & -             &                 & -               & -            & -            & -               & -               & -            & -            & -               & -               & -            \\ 
\textbf{f-Cal-KL (ours)}      & 1.42                  & 1.52              & 0.56            & 9.21            & -1.76        & 9.679          & 2.911         & 0.074           & 22.5            & 2.004        & 51.874       & 0.162           & 4.126           & 0.846        & 38.481       & 0.126           & 1.686           & 0.929        \\ 
\textbf{f-Cal-Wass (Ours)}    & 1.43                  & 1.54              & 0.79            & 7.99            & -1.75        & 9.509          & 3.202         & 0.156           & 67.9            & 2.157        & 48.04        & 0.115           & 0.768           & 0.914        & 37.220       & 0.104           & 0.832           & 1.007        \\ %
\end{tabular}%
}
\vspace{0.2cm}
\captionof{table}{\textbf{\coolname{} - Results}: We evaluate \coolname{} for a wide range of robot perception tasks and datasets. In each column group (a, b, c, d), we report an empirical risk (deterministic performance metric such as L1, SiLog, RMSE, mAP), expected calibration errors (ECE), and negative log-likelihood. \coolname{} consistently outperforms all other calibration techniques considered (lower ECE values). (a) We develop Bokeh -- a synthetic disc tracking benchmark that contains GT uncertainty values, useful for baseline comparisons. (b) Depth estimation on the KITTI benchmark~\cite{kitti}. (c) Object detection on the KITTI benchmark~\cite{kitti}. (d) Object detection on the Cityscapes dataset~\cite{cordts2016cityscapes}.
Notably, \coolname{} improves calibration without sacrificing deterministic performance. (Note: L1 scores are scaled by a factor of $1000$ and ECE scores by a factor $100$ for improved readability.
$\downarrow$: Lower is better, $\uparrow$: Higher is better, $-$: Method did not scale to task/dataset)}
\vspace{-0.4cm}
\label{table:all-results}
\end{table*}

\subsection{Evaluation metrics}
We evaluate the accuracy in calibration by means of the following widely used metrics.
The \textbf{expected calibration error} (ECE)~\cite{naeini2015obtaining, bosch-calib} measures the discrete discrepancy between the predicted distribution of the neural regressor and that of the label distribution. We divide the predicted distribution into $S$ intervals of size $\frac{1}{S}$. ECE is computed as the difference between the empirical bin frequency and the true frequency($\frac{1}{S}$). For total samples $P$ and number of samples in bin $s$ as $B_s$, $ECE =  \sum_{s = 1}^{S} \frac{|B_s|}{P} \left\lVert \frac{1}{S} - \frac{|B_s|}{P} \right\rVert $.

\begin{figure}[!ht]
    \centering
    \includegraphics[width=\columnwidth]{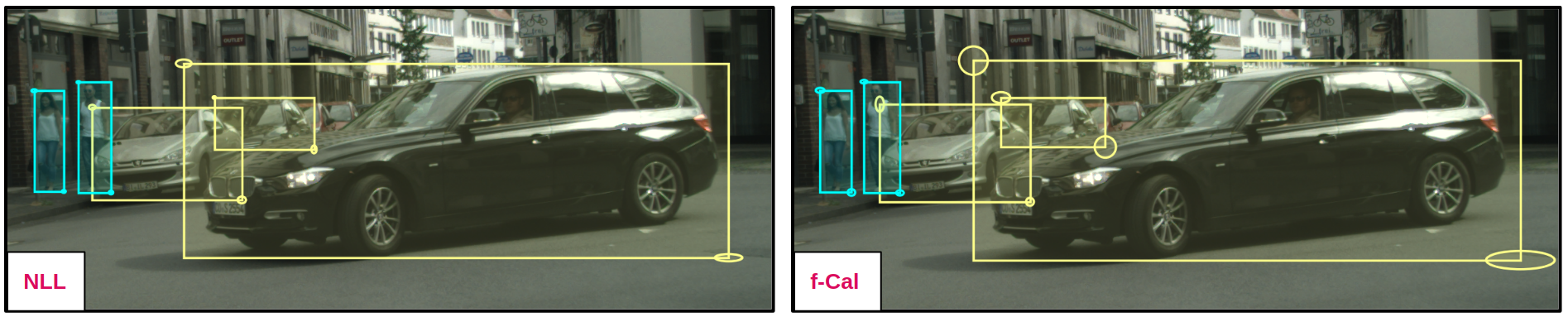}
    \vspace{-0.2cm}
    \caption{\textbf{Qualitative results}: Uncertainty calibration for object detection models (Faster RCNN) over the KITTI~\cite{kitti} dataset. (\textit{Left}) Models trained using an NLL loss term produce overconfident predictions (notice how the model outputs small, low uncertainty, ellipses for the occluded cars). (\textit{Right}) \coolname{}, on the other hand, produces calibrated uncertainty estimates (notice the large covariances for occluded cars, and the car in the foreground, whose endpoints are indeed uncertain).
    }
    \label{fig:qualitative_results}
\end{figure}

We report ECE scores for standard normal distribution and chi-squared distribution in this work, which we denote by ECE(z) and ECE(q) respectively. 
We also plot \textbf{reliability diagrams} which visually depict the amount of miscalibration over the support of the distribution. Perfectly calibrated distributions should have a diagonal reliability plot. Portions of a curve above the diagonal line are over-confident regions, while those below the curve are under-confident.

\subsection{Bokeh: A synthetic disc-tracking benchmark}
\label{subsec:bokeh}

Since ground-truth estimates of aleatoric uncertainty are extremely challenging to obtain from real-world datasets, we first validate our proposed approach in simulation.

\textbf{Setup}: We design a synthetic dataset akin to~\cite{backpropkf} for a \emph{disc-tracking} task.
The goal is to predict the 2D location of the centre of a unique red disc from an input image containing other distractor discs.
All disc locations are sampled from a known data-generating distribution.

\textbf{Models}: We use a 3-layer ConvNet architecture with an uncertainty prediciton head.
We train a model using the NLL loss~\cite{loss_att} for our baseline probabilistic regressor.
We then train two models using our proposed \coolname{} loss (\coolname{}-KL and \coolname{}-Wass).

\textbf{Results:} Table~\ref{table:all-results}(a) compares \coolname{} to the aforementioned baselines, evaluating \emph{performance} (i.e., the accuracy of the estimated mean) and \emph{calibration} quality.

\begin{figure}[!ht]
    \centering
    \includegraphics[width=\columnwidth]{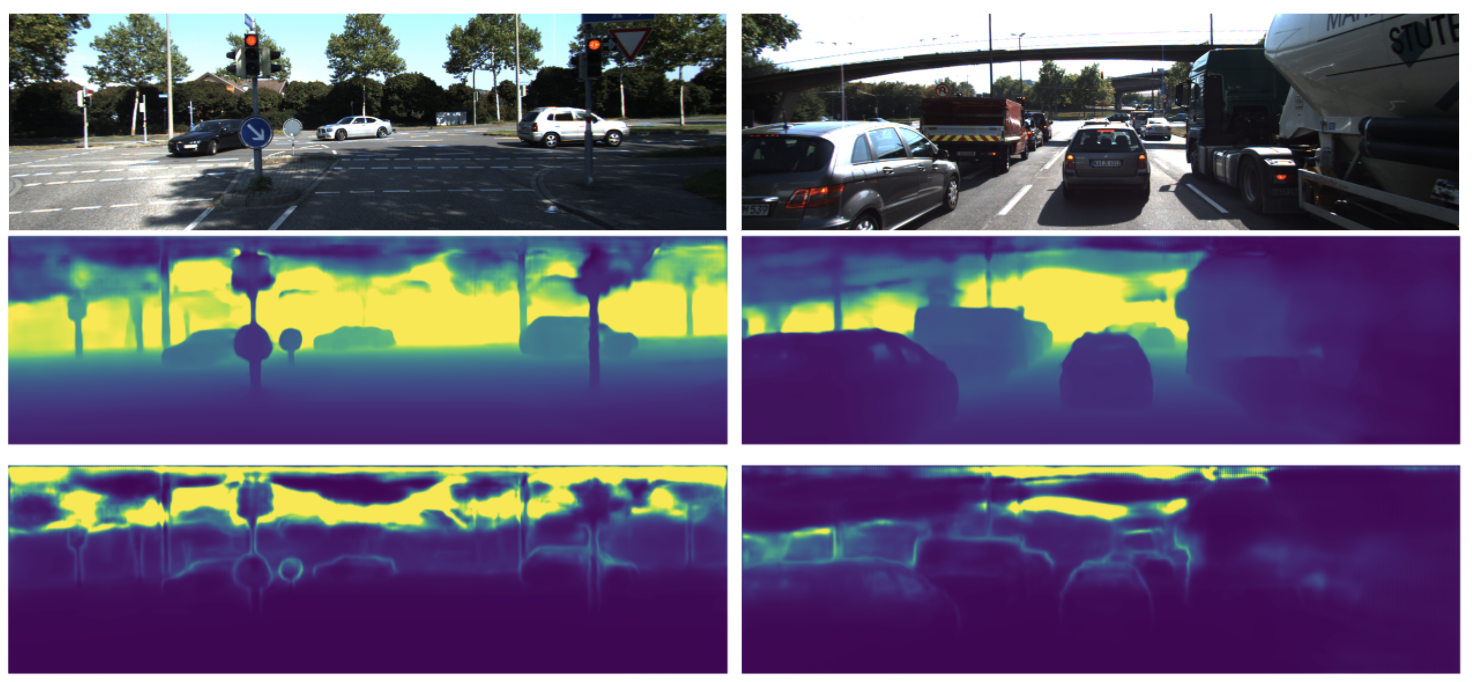}
    \vspace{0.1cm}
    \caption{\textbf{Qualitative results} for depth estimation models on the KITTI~\cite{kitti} benchmark. (\textit{Top}) Input image; (\textit{Middle}) Predicted depth; (\textit{Bottom}) Predicted uncertainty. 
    }
    \label{fig:depth_qualitative}
\end{figure}

\begin{figure}[!ht]
    \centering
    \includegraphics[width=\columnwidth]{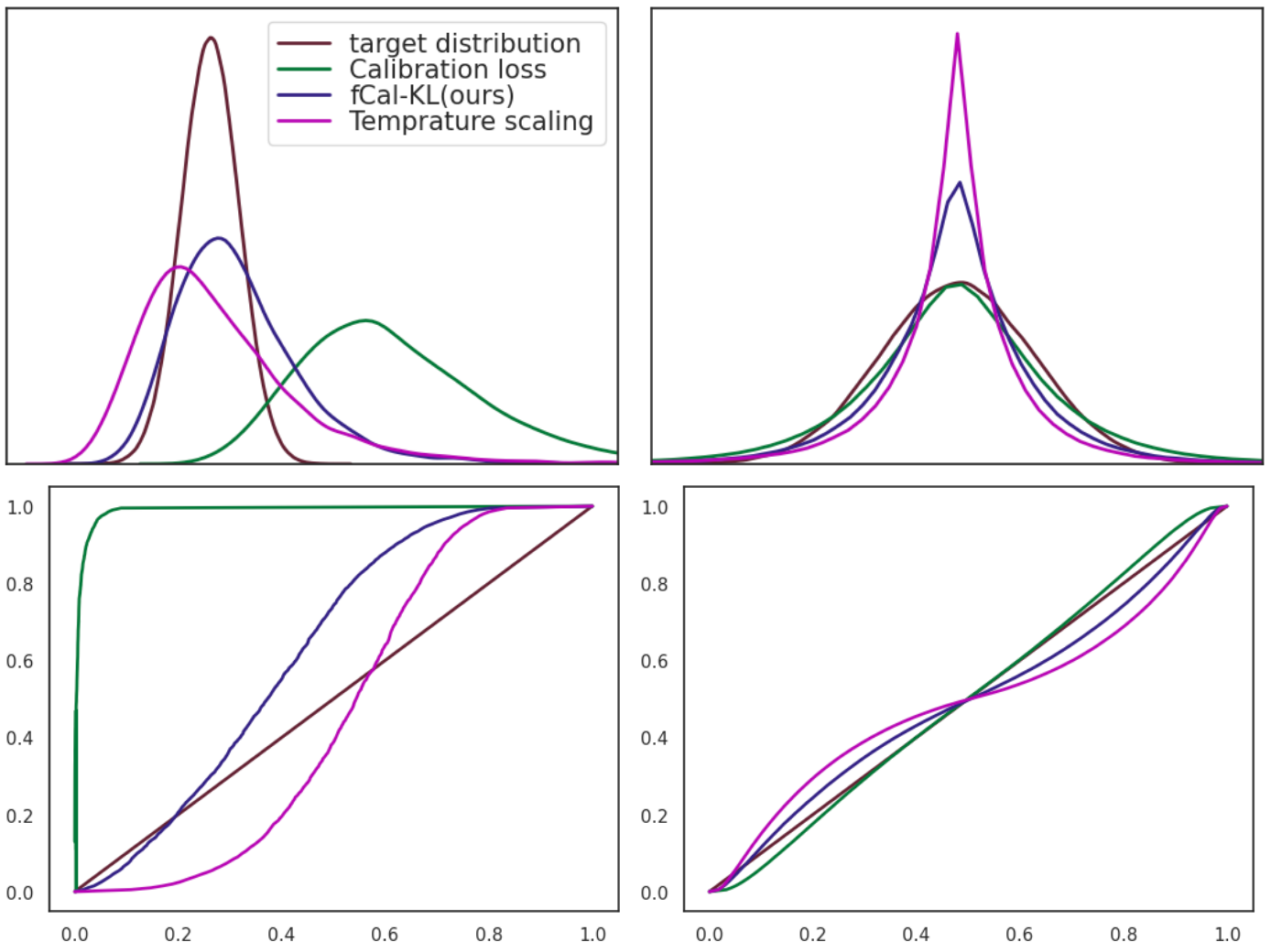}
    \vspace{0.1cm}
    \caption{\textbf{Calibration plots} on KITTI\cite{kitti} object detection: \textit{Top:} Predicted Chi-squared distributions (using hyper-constraints) and standard normal distributions from the residuals, \textit{Bottom:} corresponding reliability diagrams for chi-square and standard normal space. \coolname{} consistently yields superior calibration curves in both, chi-square and standard normal space. These curves correspond to results reported in Table \ref{table:all-results}}
    \vspace{-0.3cm}
    \label{fig:kitti-results}
\end{figure}

We report the performance (Smooth-L1 error) denoted by L1 in Table \ref{table:all-results} for both the \emph{noise-free} ground-truth (in typical ML settings, we never have access to this variable. We only ever access the noisy ground-truth labels), and the \emph{noisy} ground-truth (accounting for label generation error).

We see in Table \ref{table:all-results} that \coolname{} outperforms all baselines considered.
It is worth noting that we perform better than temperature scaling~\cite{bosch-calib} despite this being a somewhat unfair comparison (temperature scaling leverages a large held-out calibration dataset, while we do not use any additional data). \coolname{} gives well-calibrated uncertainty estimates without sacrificing the deterministic performance (more discussion of this point in Sec.~\ref{subsec:analysis}).

\subsection{KITTI Depth Estimation}
\vspace{-0.1cm}

\textbf{Setup}: We evaluate \coolname{} on real-world robotics tasks like depth estimation and object detection (Sec. \ref{subsec:object_detection}). We train \coolname{} and several baseline calibration techniques on the KITTI depth estimation benchmark dataset~\cite{kitti}. We modify the BTS model\cite{big2small} for supervised depth estimation into a Bayesian Neural Network by adding a variance decoder. We evaluate the deterministic performance using SiLog and RMSE metrics and calibration using ECE and NLL. 

\begin{figure}[tb]
    \centering
    \includegraphics[width=\columnwidth]{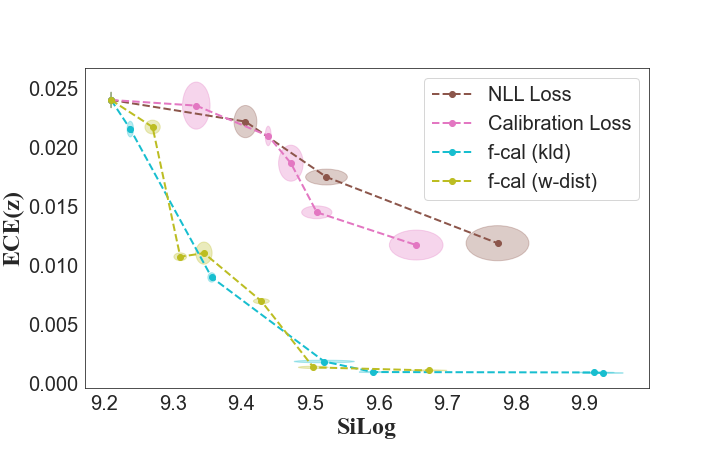}
    \vspace{-0.1cm}
    \caption{\textbf{Calibration-vs-deterministic performance trade-off:} We see that this trade-off is observed for all the three calibration techniques. For similar deterministic performance \coolname{} models are able to achieve smaller ECE values (i.e., better calibration). }
    \vspace{-0.6cm}
    \label{fig:trade-off}
\end{figure}

\textbf{Discussion}: Through our experiments, we conclude that there is a trade-off between deterministic and calibration performances as shown in Fig.~\ref{fig:trade-off} (also established in \cite{guo2017calibration, bosch-calib}). We can control this trade-off by varying the $\lambda$ in \eqref{eq:fcal}. By plotting SiLog and ECE for different values of $\lambda$ we can analyze this trade-off for the baseline calibration techniques. We note that $\lambda$ may be application dependent. To our knowledge our method is the first that enables this tradeoff to be made easily with one parameter.  
In Table.~\ref{table:all-results}-(b), for every method we select a $\lambda$ which best balances between deterministic performance and calibration. For this fixed $\lambda$ we run the experiment over multiple seeds and report mean scores 
We see that \coolname{} outperforms all baselines on all calibration metrics. We also observe that unlike Bokeh (Table. \ref{table:all-results} - (a)), temperature scaling struggles to calibrate uncertainties by tuning a single temperature parameter on such a large and complex task of depth estimation. We show qualitative results of depth estimation in figure~\ref{fig:depth_qualitative}.

\vspace{-0.15cm}
\subsection{Object detection}
\label{subsec:object_detection}
\vspace{-0.1cm}

\textbf{Setup}: We now consider the task of object detection in an autonomous driving seting. We calibrate probabilistic object detectors trained on the KITTI~\cite{kitti} and Cityscapes~\cite{cordts2016cityscapes} datasets.%
We use the popular Faster R-CNN~\cite{ren2015faster} model with a feature pyramid network~\cite{lin2017feature} and a Resnet-101~\cite{he2016deep} backbone. We use the publicly available detectron2~\cite{wu2019detectron2} implementation and extend the model to output variances.

\textbf{Discussion}: We summarize the results of our object detection experiments in Table~\ref{table:all-results}-(c, d) and Fig.~\ref{fig:kitti-results}. As can be seen in Table~\ref{table:all-results}, we see that \coolname{} variants, while having competitive regression performance (in terms of mAP), exhibit far superior calibration as reflected through ECE scores.
In Fig~\ref{fig:kitti-results}, we can see through reliability plots that the baselines methods yield inferior calibration and are farther away from the ground-truth distribution. It is important to note that even though calibration loss (\cite{bosch-calib}) is able  to predict a distribution which is close to being standard normal, it is still not as calibrated as the \coolname{} estimates. This is reflected in the  reliability diagram for the Chi-squared distribution which is much more contrastive than the curve for the standard normal distribution. Fig~\ref{fig:kitti-results} also shows that loss attenuation yields very over-confident uncertainty predictions, which can be corroborated with qualitative results shown in Figure \ref{fig:qualitative_results}. By employing hyper-constraints over the proposed distribution, \coolname{} enforces regularization at a batch level which leads to superior calibration performance.

\section{Discussion and Conclusion}
\label{sec:limitations}

\begin{figure}[tb]
    \centering
    \includegraphics[width=\columnwidth]{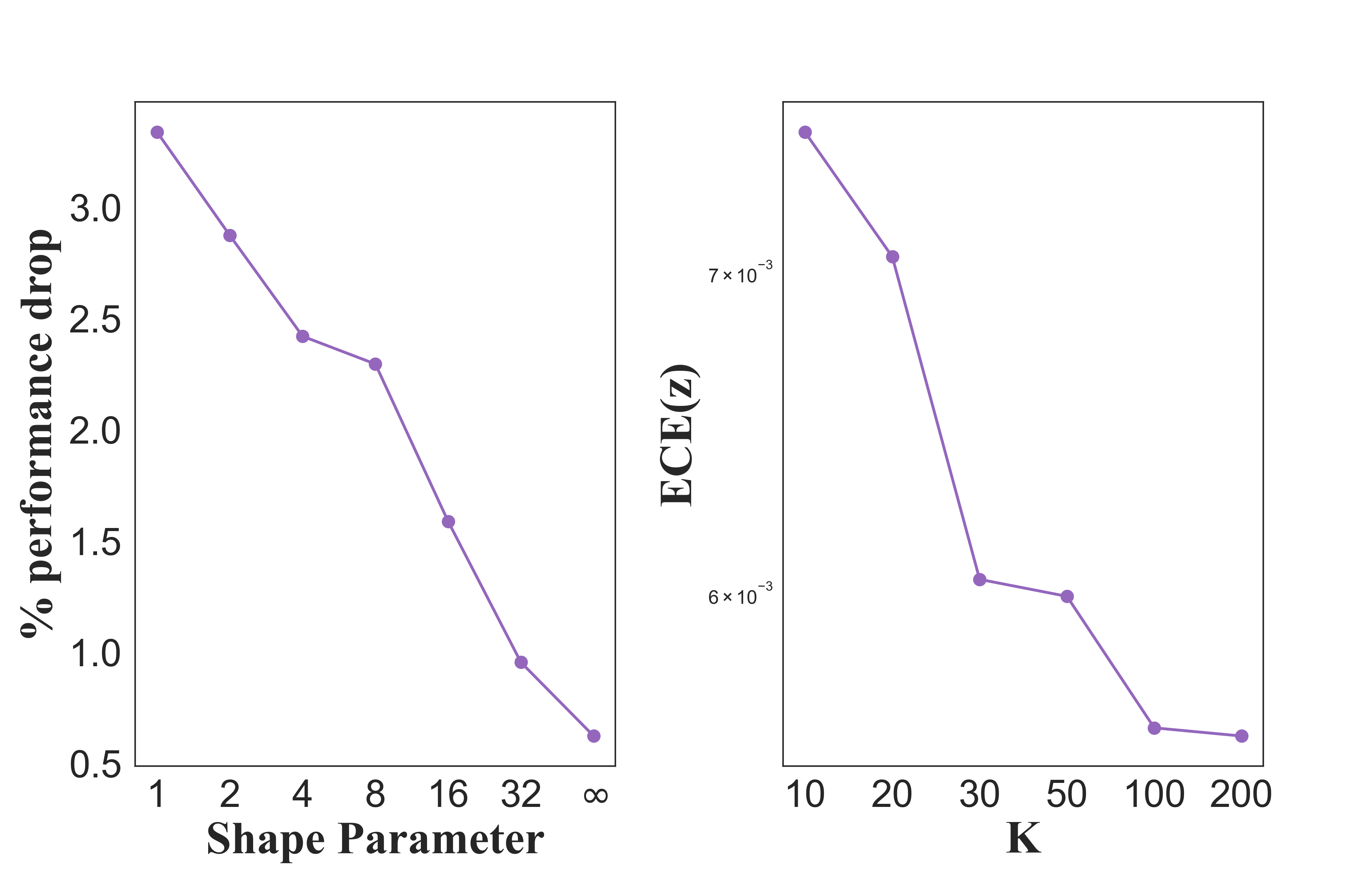}
    \caption{\textbf{Ablation}: (left) We plot the \% drop in deterministic performance compared to a deterministic model for different noise distributions. For large shape parameter, the Gamma distribution converges to a Gaussian, resulting in nearly identical performance to a deterministic model. (right) Effect of K on the performance of \coolname{}, we see that as long as $K > 50$, the central limit theorem holds and we get good calibration. }
    \label{fig:ablation}
\end{figure}

\textbf{Impact of modeling assumption}:
\label{subsec:analysis}
We postulate that for real-world datasets such as KITTI \cite{kitti}, the tradeoff in calibration and deterministic performance occurs due to poor modeling assumptions (i.e., modeling uncertainty using a distribution that is quite different from the underlying label error distribution).
To investigate this, we introduce a mismatch between the true distribution, a Gamma distribution parameterized by $\gamma$, and the assumed distribution,  a Gaussian distribution, on the synthetic (Bokeh) dataset (Fig. \ref{fig:ablation} (left)). For lower distributional mismatch, the performance gap between the calibrated and deterministic models is reduced. We attribute the deterministic performance drop for KITTI results to this phenomenon. 

The impact of this facet of our approach is significant. This means that through experimenting with different modeling assumptions and looking at the resulting tradeoff, we may be able to infer something about the underlying noise distribution, something that is typically very hard to do.

\textbf{Effect of degrees of freedom (K)}:
We analyze how the number of degrees of freedom ($K$) would impact calibration performance. We train models with different values of $K$ and measure the degree of calibration. In Fig. \ref{fig:ablation} (right), we can observe that for $K > $  50, the central limit theorem holds and we see superior calibration when compared with models trained for $K \leq $  50, when our approximation of a Gaussian distribution breaks, resulting in poor calibration. For Object detection(where thousands of proposals are being scored) and per pixel depth estimation, minibatch size $N >> K$, which allows us to effectively construct hyperconstraints.

\textbf{Summary}: In this work, we presented \coolname{}, a principled variational inference approach to calibrate aleatoric uncertainty estimates from deep neural networks.
This enables the deep neural network perceptual models to be treated as a sensor in a typical robot autonomy stack. Uncertainties associated with object detectors can be employed in object-based state estimation or in model-predictive control loops. In future, we will extend the approach to also consider epistemic uncertainty estimation. Another interesting avenue for future work could be to investigate non-iid settings -- common in sequential and online learning scenarios.

\bibliographystyle{IEEEtranS}
\bibliography{IEEEabrv,references/uncertainty_calibration}  %

\begin{thebibliography}{10}
\providecommand{\url}[1]{#1}
\csname url@rmstyle\endcsname
\providecommand{\newblock}{\relax}
\providecommand{\bibinfo}[2]{#2}
\providecommand\BIBentrySTDinterwordspacing{\spaceskip=0pt\relax}
\providecommand\BIBentryALTinterwordstretchfactor{4}
\providecommand\BIBentryALTinterwordspacing{\spaceskip=\fontdimen2\font plus
\BIBentryALTinterwordstretchfactor\fontdimen3\font minus
  \fontdimen4\font\relax}
\providecommand\BIBforeignlanguage[2]{{%
\expandafter\ifx\csname l@#1\endcsname\relax
\typeout{** WARNING: IEEEtran.bst: No hyphenation pattern has been}%
\typeout{** loaded for the language `#1'. Using the pattern for}%
\typeout{** the default language instead.}%
\else
\language=\csname l@#1\endcsname
\fi
#2}}

\bibitem{romandhaivat}
D.~Bhatt, A.~Garg, B.~Gopalakrishnan, and K.~M. Krishna, ``Probabilistic
  obstacle avoidance and object following: An overlap of gaussians approach,''
  in \emph{2019 28th IEEE International Conference on Robot and Human
  Interactive Communication (RO-MAN)}, 2019, pp. 1--8.

\bibitem{blackmore2006probabilistic}
L.~Blackmore, H.~Li, and B.~Williams, ``A probabilistic approach to optimal
  robust path planning with obstacles,'' in \emph{American Control Conference,
  2006}.\hskip 1em plus 0.5em minus 0.4em\relax IEEE, 2006, pp. 7--pp.

\bibitem{blundell2015weight}
C.~Blundell, J.~Cornebise, K.~Kavukcuoglu, and D.~Wierstra, ``Weight
  uncertainty in neural network,'' in \emph{International Conference on Machine
  Learning}.\hskip 1em plus 0.5em minus 0.4em\relax PMLR, 2015, pp. 1613--1622.

\bibitem{chung2020beyond}
Y.~Chung, W.~Neiswanger, I.~Char, and J.~Schneider, ``Beyond pinball loss:
  Quantile methods for calibrated uncertainty quantification,'' \emph{arXiv
  preprint arXiv:2011.09588}, 2020.

\bibitem{cordts2016cityscapes}
M.~Cordts, M.~Omran, S.~Ramos, T.~Rehfeld, M.~Enzweiler, R.~Benenson,
  U.~Franke, S.~Roth, and B.~Schiele, ``The cityscapes dataset for semantic
  urban scene understanding,'' in \emph{Proceedings of the IEEE conference on
  computer vision and pattern recognition}, 2016, pp. 3213--3223.

\bibitem{cui2020calibrated}
P.~Cui, W.~Hu, and J.~Zhu, ``Calibrated reliable regression using maximum mean
  discrepancy,'' \emph{Advances in Neural Information Processing Systems},
  vol.~33, 2020.

\bibitem{embrechts2013note}
P.~Embrechts and M.~Hofert, ``A note on generalized inverses,''
  \emph{Mathematical Methods of Operations Research}, vol.~77, no.~3, pp.
  423--432, 2013.

\bibitem{bosch-calib}
D.~Feng, L.~Rosenbaum, C.~Glaeser, F.~Timm, and K.~Dietmayer, ``Can we trust
  you? on calibration of a probabilistic object detector for autonomous
  driving,'' \emph{arXiv preprint arXiv:1909.12358}, 2019.

\bibitem{gal2016uncertainty}
Y.~Gal, ``Uncertainty in deep learning,'' \emph{University of Cambridge},
  vol.~1, no.~3, 2016.

\bibitem{gal2016dropout}
Y.~Gal and Z.~Ghahramani, ``Dropout as a bayesian approximation: Representing
  model uncertainty in deep learning,'' in \emph{international conference on
  machine learning}.\hskip 1em plus 0.5em minus 0.4em\relax PMLR, 2016, pp.
  1050--1059.

\bibitem{gast2018lightweight}
J.~Gast and S.~Roth, ``Lightweight probabilistic deep networks,'' in
  \emph{Proceedings of the IEEE Conference on Computer Vision and Pattern
  Recognition}, 2018, pp. 3369--3378.

\bibitem{kitti}
A.~Geiger, P.~Lenz, and R.~Urtasun, ``Are we ready for autonomous driving? the
  kitti vision benchmark suite,'' in \emph{2012 IEEE Conference on Computer
  Vision and Pattern Recognition}.\hskip 1em plus 0.5em minus 0.4em\relax IEEE,
  2012, pp. 3354--3361.

\bibitem{gopalakrishnan2017prvo}
B.~Gopalakrishnan, A.~K. Singh, M.~Kaushik, K.~M. Krishna, and D.~Manocha,
  ``Prvo: Probabilistic reciprocal velocity obstacle for multi robot navigation
  under uncertainty,'' in \emph{2017 IEEE/RSJ International Conference on
  Intelligent Robots and Systems (IROS)}.\hskip 1em plus 0.5em minus
  0.4em\relax IEEE, 2017, pp. 1089--1096.

\bibitem{guo2017calibration}
C.~Guo, G.~Pleiss, Y.~Sun, and K.~Q. Weinberger, ``On calibration of modern
  neural networks,'' in \emph{International Conference on Machine
  Learning}.\hskip 1em plus 0.5em minus 0.4em\relax PMLR, 2017, pp. 1321--1330.

\bibitem{backpropkf}
T.~Haarnoja, A.~Ajay, S.~Levine, and P.~Abbeel, ``Backprop kf: Learning
  discriminative deterministic state estimators,'' in \emph{Advances in Neural
  Information Processing Systems}, 2016, pp. 4376--4384.

\bibitem{hall2020probabilistic}
D.~Hall, F.~Dayoub, J.~Skinner, H.~Zhang, D.~Miller, P.~Corke, G.~Carneiro,
  A.~Angelova, and N.~S{\"u}nderhauf, ``Probabilistic object detection:
  Definition and evaluation,'' in \emph{Proceedings of the IEEE/CVF Winter
  Conference on Applications of Computer Vision}, 2020, pp. 1031--1040.

\bibitem{harris2020array}
\BIBentryALTinterwordspacing
C.~R. Harris, K.~J. Millman, S.~J. van~der Walt, R.~Gommers, P.~Virtanen,
  D.~Cournapeau, E.~Wieser, J.~Taylor, S.~Berg, N.~J. Smith, R.~Kern, M.~Picus,
  S.~Hoyer, M.~H. van Kerkwijk, M.~Brett, A.~Haldane, J.~F. del R{\'{i}}o,
  M.~Wiebe, P.~Peterson, P.~G{\'{e}}rard-Marchant, K.~Sheppard, T.~Reddy,
  W.~Weckesser, H.~Abbasi, C.~Gohlke, and T.~E. Oliphant, ``Array programming
  with {NumPy},'' \emph{Nature}, vol. 585, no. 7825, pp. 357--362, Sept. 2020.
  [Online]. Available: \url{https://doi.org/10.1038/s41586-020-2649-2}
\BIBentrySTDinterwordspacing

\bibitem{he2016deep}
K.~He, X.~Zhang, S.~Ren, and J.~Sun, ``Deep residual learning for image
  recognition,'' in \emph{Proceedings of the IEEE conference on computer vision
  and pattern recognition}, 2016, pp. 770--778.

\bibitem{he2019bounding}
Y.~He, C.~Zhu, J.~Wang, M.~Savvides, and X.~Zhang, ``Bounding box regression
  with uncertainty for accurate object detection,'' in \emph{Proceedings of the
  IEEE Conference on Computer Vision and Pattern Recognition}, 2019, pp.
  2888--2897.

\bibitem{hein2019relu}
M.~Hein, M.~Andriushchenko, and J.~Bitterwolf, ``Why relu networks yield
  high-confidence predictions far away from the training data and how to
  mitigate the problem,'' in \emph{Proceedings of the IEEE Conference on
  Computer Vision and Pattern Recognition}, 2019, pp. 41--50.

\bibitem{hendrycks2016baseline}
D.~Hendrycks and K.~Gimpel, ``A baseline for detecting misclassified and
  out-of-distribution examples in neural networks,'' \emph{arXiv preprint
  arXiv:1610.02136}, 2016.

\bibitem{hendrycks2018deep}
D.~Hendrycks, M.~Mazeika, and T.~Dietterich, ``Deep anomaly detection with
  outlier exposure,'' in \emph{International Conference on Learning
  Representations}, 2018.

\bibitem{ho2005calibrated}
Y.~H. Ho and S.~M. Lee, ``Calibrated interpolated confidence intervals for
  population quantiles,'' \emph{Biometrika}, vol.~92, no.~1, pp. 234--241,
  2005.

\bibitem{kendall2015bayesian}
A.~Kendall, V.~Badrinarayanan, and R.~Cipolla, ``Bayesian segnet: Model
  uncertainty in deep convolutional encoder-decoder architectures for scene
  understanding,'' \emph{arXiv preprint arXiv:1511.02680}, 2015.

\bibitem{kendall2017uncertainties}
A.~Kendall and Y.~Gal, ``What uncertainties do we need in bayesian deep
  learning for computer vision?'' in \emph{Advances in neural information
  processing systems}, 2017, pp. 5574--5584.

\bibitem{kuleshov2018accurate}
V.~Kuleshov, N.~Fenner, and S.~Ermon, ``Accurate uncertainties for deep
  learning using calibrated regression,'' in \emph{International Conference on
  Machine Learning}.\hskip 1em plus 0.5em minus 0.4em\relax PMLR, 2018, pp.
  2796--2804.

\bibitem{lakshminarayanan2017simple}
B.~Lakshminarayanan, A.~Pritzel, and C.~Blundell, ``Simple and scalable
  predictive uncertainty estimation using deep ensembles,'' \emph{Advances in
  Neural Information Processing Systems}, vol.~30, 2017.

\bibitem{big2small}
J.~H. Lee, M.-K. Han, D.~W. Ko, and I.~H. Suh, ``From big to small: Multi-scale
  local planar guidance for monocular depth estimation,'' \emph{arXiv preprint
  arXiv:1907.10326}, 2019.

\bibitem{levi2019evaluating}
D.~Levi, L.~Gispan, N.~Giladi, and E.~Fetaya, ``Evaluating and calibrating
  uncertainty prediction in regression tasks,'' \emph{arXiv preprint
  arXiv:1905.11659}, 2019.

\bibitem{liang2017enhancing}
S.~Liang, Y.~Li, and R.~Srikant, ``Enhancing the reliability of
  out-of-distribution image detection in neural networks,'' in
  \emph{International Conference on Learning Representations}, 2018.

\bibitem{lin2017feature}
T.-Y. Lin, P.~Doll{\'a}r, R.~Girshick, K.~He, B.~Hariharan, and S.~Belongie,
  ``Feature pyramid networks for object detection,'' in \emph{Proceedings of
  the IEEE conference on computer vision and pattern recognition}, 2017, pp.
  2117--2125.

\bibitem{lin2014microsoft}
T.-Y. Lin, M.~Maire, S.~Belongie, J.~Hays, P.~Perona, D.~Ramanan,
  P.~Doll{\'a}r, and C.~L. Zitnick, ``Microsoft coco: Common objects in
  context,'' in \emph{European conference on computer vision}.\hskip 1em plus
  0.5em minus 0.4em\relax Springer, 2014, pp. 740--755.

\bibitem{malinin2018predictive}
A.~Malinin and M.~Gales, ``Predictive uncertainty estimation via prior
  networks,'' in \emph{Advances in Neural Information Processing Systems},
  2018, pp. 7047--7058.

\bibitem{mohseni2020self}
S.~Mohseni, M.~Pitale, J.~Yadawa, and Z.~Wang, ``Self-supervised learning for
  generalizable out-of-distribution detection,'' in \emph{Proceedings of the
  AAAI Conference on Artificial Intelligence}, vol.~34, no.~04, 2020, pp.
  5216--5223.

\bibitem{naeini2015obtaining}
M.~P. Naeini, G.~F. Cooper, and M.~Hauskrecht, ``Obtaining well calibrated
  probabilities using bayesian binning,'' in \emph{Proceedings of the... AAAI
  Conference on Artificial Intelligence. AAAI Conference on Artificial
  Intelligence}, vol. 2015.\hskip 1em plus 0.5em minus 0.4em\relax NIH Public
  Access, 2015, p. 2901.

\bibitem{loss_att}
D.~A. Nix and A.~S. Weigend, ``Estimating the mean and variance of the target
  probability distribution,'' in \emph{Proceedings of 1994 ieee international
  conference on neural networks (ICNN'94)}, vol.~1.\hskip 1em plus 0.5em minus
  0.4em\relax IEEE, 1994, pp. 55--60.

\bibitem{paszke2019pytorch}
A.~Paszke, S.~Gross, F.~Massa, A.~Lerer, J.~Bradbury, G.~Chanan, T.~Killeen,
  Z.~Lin, N.~Gimelshein, L.~Antiga, \emph{et~al.}, ``Pytorch: An imperative
  style, high-performance deep learning library,'' in \emph{Advances in neural
  information processing systems}, 2019, pp. 8026--8037.

\bibitem{ren2015faster}
S.~Ren, K.~He, R.~Girshick, and J.~Sun, ``Faster r-cnn: Towards real-time
  object detection with region proposal networks,'' in \emph{Advances in neural
  information processing systems}, 2015, pp. 91--99.

\bibitem{rueda2007calibration}
M.~Rueda, S.~Mart{\'\i}nez-Puertas, H.~Mart{\'\i}nez-Puertas, and A.~Arcos,
  ``Calibration methods for estimating quantiles,'' \emph{Metrika}, vol.~66,
  no.~3, pp. 355--371, 2007.

\bibitem{sehwag2019analyzing}
V.~Sehwag, A.~N. Bhagoji, L.~Song, C.~Sitawarin, D.~Cullina, M.~Chiang, and
  P.~Mittal, ``Analyzing the robustness of open-world machine learning,'' in
  \emph{Proceedings of the 12th ACM Workshop on Artificial Intelligence and
  Security}, 2019, pp. 105--116.

\bibitem{shin2018direct}
Y.-S. Shin, Y.~S. Park, and A.~Kim, ``Direct visual slam using sparse depth for
  camera-lidar system,'' in \emph{2018 IEEE International Conference on
  Robotics and Automation (ICRA)}.\hskip 1em plus 0.5em minus 0.4em\relax IEEE,
  2018, pp. 5144--5151.

\bibitem{gp-beta}
H.~Song, T.~Diethe, M.~Kull, and P.~Flach, ``Distribution calibration for
  regression,'' in \emph{International Conference on Machine Learning}.\hskip
  1em plus 0.5em minus 0.4em\relax PMLR, 2019, pp. 5897--5906.

\bibitem{tagasovska2019single}
N.~Tagasovska and D.~Lopez-Paz, ``Single-model uncertainties for deep
  learning,'' in \emph{Advances in Neural Information Processing Systems},
  2019, pp. 6417--6428.

\bibitem{taillardat2016calibrated}
M.~Taillardat, O.~Mestre, M.~Zamo, and P.~Naveau, ``Calibrated ensemble
  forecasts using quantile regression forests and ensemble model output
  statistics,'' \emph{Monthly Weather Review}, vol. 144, no.~6, pp. 2375--2393,
  2016.

\bibitem{wu2019detectron2}
Y.~Wu, A.~Kirillov, F.~Massa, W.-Y. Lo, and R.~Girshick, ``Detectron2,''
  \url{https://github.com/facebookresearch/detectron2}, 2019.

\bibitem{isotonic}
B.~Zadrozny and C.~Elkan, ``Transforming classifier scores into accurate
  multiclass probability estimates,'' in \emph{Proceedings of the eighth ACM
  SIGKDD international conference on Knowledge discovery and data mining},
  2002, pp. 694--699.

\end{thebibliography}

\newpage
\onecolumn
{\Large \appendix}

\section{Deriving \coolname{} loss:}
In this section, we derive the KL-divergence and W-dist loss presented in this work. 

Here, let's say that the neural regressor is predicting N regression variables over an entire batch of inputs. We use the following notations for our predictions and ground-truth. 
\begin{itemize}
    \item predicted means: $\mu_1, \mu_2, ...., \mu_N$
    \item predicted variance: $\sigma_1^2, \sigma_2^2, ...., \sigma_N^2$
    \item Ground truth: $y_1, y_2, ...., y_N$
    \item K = degrees of freedom of a chi-squared random variable, generally, K $ > $ 50. 
\end{itemize}
Here, N is assumed to be larger, generally N $>$ 1000. Here K is a hyper-parameters. 
\begin{itemize}
    \item $z_i^2 = \frac{(y_i - \mu_i)^2}{\sigma_i^2} \sim \chi_1^2 $; i = \{1, 2, ..., N\}, are Mahalanobis distances with DoF 1.
\end{itemize}
\begin{align*}
    & Q_i = \sum_{j = 1}^{K} z_{ij}^2 = \sum_{j = 1}^{K} \frac{(y_{ij} - \mu_{ij})^2}{\sigma_{ij}^2} \\
    & y_{ij} {\sim} \{y_1,y_2, ...., y_N\}  \\
    & \boxed{Q_i = \sum_{j = 1}^{K} \frac{(y_{ij} - \mu_{ij})^2}{\sigma_{ij}^2} \sim \chi_K^2}
\end{align*}

Here, $\mu_{ij}$ and $\sigma_{ij}$ are predictions corresponding to  $y_{ij}$. $y_{ij}$ is uniformly sampled without replacement. We have H such $Q_i$, each distributed as chi-squared random variable with DoF K. H is a hyper-parameter, which is number of chi-squared samples. 

The distribution resulting out of these H random variables is a chi-squared distribution. For K $>$ 50, $\chi_K^2 \to \mathcal{N}(K, 2K)$. 
Empirical mean($\mu_{\chi_K^2}$) and variance($\sigma_{\chi_K^2}^2$) of the chi-squared distribution can be written as below,

\begin{align*}
  & \mu_{\chi_K^2} = \frac{1}{H} \sum_{i = 1}^{H} Q_i = \sum_{i = 1}^H \sum_{j = 1}^{K} \frac{(y_{ij} - \mu_{ij})^2}{\sigma_{ij}^2} , \sigma_{\chi_K^2}^2 = \frac{1}{H-1} \sum_{i = 1}^{H}(Q_i - \mu_{\chi_K^2})^2 \\ 
  & \sigma_{\chi_K^2}^2 = \frac{1}{H-1} \sum_{i = 1}^{H} \bigg( \sum_{j = 1}^{K} \frac{(y_{ij} - \mu_{ij})^2}{\sigma_{ij}^2} -  \frac{1}{K}\sum_{i = 1}^H \sum_{j = 1}^{K} \frac{(y_{ij} - \mu_{ij})^2}{\sigma_{ij}^2} \bigg)^2 
\end{align*}

In the above equation, we get empirical means and variance of our chi-squared distribution. 

According to central limit theorem, Chi-squared distribution with degrees of freedom K(K $\geq$ 50) follow Gaussian distribution mean K and variance 2K. hence target mean($\hat{\mu}_{\chi_K^2}$) and target variance($\hat{\sigma}_{\chi_K^2}^2$) are, 
\begin{align*}
    \hat{\mu}_{\chi_K^2} = K, 
    \hat{\sigma}_{\chi_K^2}^2 = 2K
\end{align*}

Proposal distribution: $ \boxed{p(x) = \mathcal{N}( \hat{\mu}_{\chi_K^2}, \hat{\sigma}_{\chi_K^2}^2)} $

Target distribution: $ \boxed{ q(x) = \mathcal{N}(\mu_{\chi_K^2}, \sigma_{\chi_K^2}^2) } $ \\

We have statistics of proposal distribution($p(x)$) and target distribution($q(x)$). The closed form KL-divergence and Wasserstein distance between two univariate normal distributions can be expressed as below

\begin{equation*}
    \begin{aligned}
    \textnormal{KLD} & = KL(p||q) = \frac{1}{2} \log \bigg( \frac{\hat{\sigma}_{\chi_K^2}^2}{\sigma_{\chi_K^2}^2} \bigg) + \frac{\sigma_{\chi_K^2}^2 + (\mu_{\chi_K^2} -  \hat{\mu}_{\chi_K^2})^2}{2\hat{\sigma}_{\chi_K^2}^2} - \frac{1}{2} \\
    \end{aligned} \\
\end{equation*}

\begin{equation*}
    \boxed{
    \begin{aligned}
    \textnormal{KLD} = KL(p||q) = & \frac{1}{2} \log \Bigg( \frac{2K}{\frac{1}{H-1} \sum_{i = 1}^{H} \bigg( \sum_{j = 1}^{K} \frac{(y_{ij} - \mu_{ij})^2}{\sigma_{ij}^2} -  \frac{1}{K}\sum_{i = 1}^H \sum_{j = 1}^{K} \frac{(y_{ij} - \mu_{ij})^2}{\sigma_{ij}^2} \bigg)^2} \Bigg) + \\ 
    & \hspace{3cm} \bigg(\frac{2K + (\sum_{i = 1}^H \sum_{j = 1}^{K} \frac{(y_{ij} - \mu_{ij})^2}{\sigma_{ij}^2} -  K)^2 }{4K}\bigg) - \frac{1}{2} \\
    \end{aligned}
    }
\end{equation*}

\begin{equation*}
    \begin{aligned}
        \textnormal{W-dist} & = W(p,q) = (\mu_{\chi_K^2} -  \hat{\mu}_{\chi_K^2})^2 + (\hat{\sigma}_{\chi_K^2}^2 + \sigma_{\chi_K^2}^2 - 2\sigma_{\chi_K^2}\hat{\sigma}_{\chi_K^2}) \\
    \end{aligned} \\
\end{equation*}

\begin{equation*}
    \boxed{
    \begin{aligned}
        \textnormal{W-dist} & = W(p,q) = (\sum_{i = 1}^H \sum_{j = 1}^{K} \frac{(y_{ij} - \mu_{ij})^2}{\sigma_{ij}^2} -  K)^2 + \\
        & (2K + \frac{1}{H-1} \sum_{i = 1}^{H} \bigg( \sum_{j = 1}^{K} \frac{(y_{ij} - \mu_{ij})^2}{\sigma_{ij}^2} -  \frac{1}{K}\sum_{i = 1}^H \sum_{j = 1}^{K} \frac{(y_{ij} - \mu_{ij})^2}{\sigma_{ij}^2} \bigg)^2 -  \\ & 2\sqrt{\frac{1}{H-1} \sum_{i = 1}^{H} \bigg( \sum_{j = 1}^{K} \frac{(y_{ij} - \mu_{ij})^2}{\sigma_{ij}^2} -  \frac{1}{K}\sum_{i = 1}^H \sum_{j = 1}^{K} \frac{(y_{ij} - \mu_{ij})^2}{\sigma_{ij}^2} \bigg)^2}\sqrt{2K}) 
    \end{aligned} \\
    }
\end{equation*}

\section{Implementation Details}

\subsection{Bokeh: A synthetic disc-tracking benchmark}

\textbf{Dataset}: We design Bokeh with complexities of a typical regression problem for computer vision task in mind. 
The goal is to predict the 2D location of the centre of a red disc from an input image containing other distractor discs.
All discs are sampled from a known data-generating distribution\footnote{This is important, as it devoids us of the usual handicaps with real data, where one may not have access to the label-generating distribution.}.
Randomly coloured discs are added to the image to occlude the red disc and act as distractors. The locations and radii of these distractor discs are sampled from a uniform distribution. We generate homoscedastic and heteroscedastic variants of the dataset. 

We introduce noise to our ground-truth labels and create two separate synthetic datasets one where noise is Homoscedastic and other where its Heteroscedastic. Noise in x and y are sampled independently from a gaussian distribution. In case of homoscedastic noise, the noise generating distribution is $\mathcal N(0, \sigma)$, where $\sigma$ is a fixed value. On the other hand, heteroscedastic noise is generated from the distribution $\mathcal N(0, \sigma(x))$, where $\sigma$ is a function of the input image $x$. $\sigma(x)$ depends on the proximity of the distractor discs in relation to the red disc. Simply put, if the distractor discs are nearby or occluding the red disc the $\sigma(x)$ value will be high and low when they are far away. We split the dataset into training, validation and test sets in proportion of 3:1:1.

\textbf{Training:}

We train \coolname{}-KLD and \coolname{}-Wass with KL-divergence and Wasserstein distance as our loss function to measure the distance between predicted and groundtruth chi-squared distribution. All the baseline calibration  methods (Table~\ref{table:toy_exp_full}) were initialized with NLL loss trained weights.

\begin{figure*}[ht]
    \centering
    \includegraphics[width=\columnwidth]{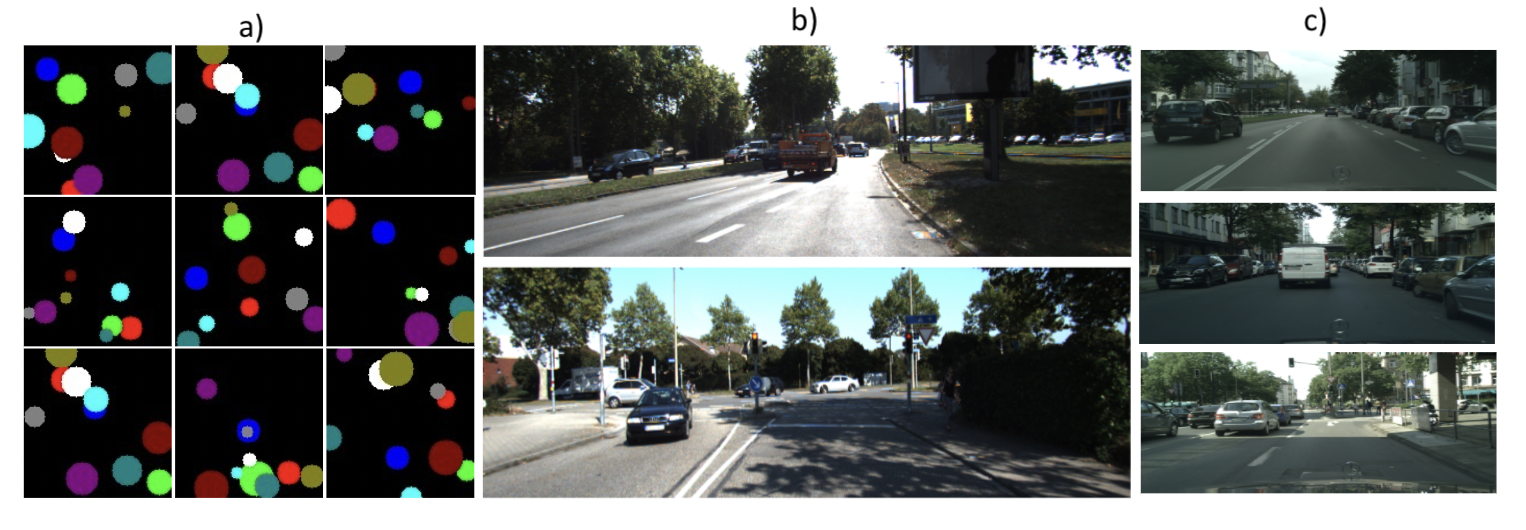}
    \caption{\textbf{Datasets}: We evaluate baselines, current-art, and \coolname{} on $3$ datasets and multiple tasks. a) We create a \textbf{synthetic dataset (Bokeh)} where the task is to regress the coordinates of the center of a unique red disk. b) \textbf{KITTI Object Detection} and \textbf{KITTI Depth Estimation} benchmark datasets~\cite{kitti} c) Object detection on \textbf{Cityscapes}~\cite{cordts2016cityscapes}}
    \label{fig:dataset}
\end{figure*}
\subsection{KITTI Depth Estimation}
To test the scalability of \coolname{} to real-world robotics tasks, we evaluate it on Depth estimation and Object Detection tasks. For depth estimation, we use the KITTI Depth Estimation benchmark dataset for evaluation. We modify the BTS\cite{big2small} model into a Bayesian Neural Network by adding an uncertainty decoder. Similar to BTS\cite{big2small} work, we train our network on a subset of nearly 26K images from KITTI\cite{kitti}, corresponding to different scenes not part of the test set containing 697 images. The depth maps have an upper bound of 80 meters. We include an uncertainty head which predicts the standard deviation corresponding to the output distribution. All the models were initialized with NLL loss trained weights.  

The loss function used for training these models is given in Eq. \ref{eq:depth_loss}.
\begin{equation}
    \mathcal{L} = \mathcal{L}_{reg} + \lambda*\mathcal{L}_{cal}
    \label{eq:depth_loss}
\end{equation}
Here $\mathcal{L}_{reg}$ is the SiLog loss function used in BTS\cite{big2small} paper, and $\mathcal{L}_{cal}$ can be NLL, calibration or the f-cal losses. As mentioned in the main paper, $\lambda$ is trade-off parameter which can be used to control the trade-off between calibration and deterministic performances.

\subsection{Object Detection}

 We use the popular Faster R-CNN~\cite{ren2015faster} with a feature pyramid network~\cite{lin2017feature} and a Resnet-101~\cite{he2016deep}backbone. We use the publicly available detectron2~\cite{wu2019detectron2} and PyTorch\cite{paszke2019pytorch} implementation and extend the model to regress uncertainty estimates. 
For uncertainty estimation in object detectors, we add uncertainty head in stage-2 of the network. We employ \textit{xyxy} bounding box parameterization as used in~\cite{he2019bounding} as opposed to \textit{xywh} used in~\cite{ren2015faster}. Using \textit{xyxy} bounding box parameterization ensures that we have all linear transformations over our predictions to get final bounding box. The reason for employing \textit{xyxy} bounding box parameterization is to ensure that our final prediction over the bounding box is Gaussian distribution. A Gaussian uncertainty going through non-linear transformation will lose Gaussian-ness of the prediction.  In uncertainty head, we have a fully connected layer followed by a Generalized Sigmoid non-linearity, $g(x) = \alpha + \frac{\beta - \alpha}{1 + \exp{(-\eta x)}}$. Here, $\beta$ is upper asymptote, $\alpha$ is a lower asymptote and $\eta$ is the sharpness. For all the experiments in this section, we have $\alpha = 0, \beta = 50 $ and $\eta = 0.15$. These hyperparameters are chosen to have wide range of uncertainty estimates. Using Generalized Sigmoid function bounds our variance predictions as well as provides stable training dynamics. All the baseline calibration models were initialized with NLL Loss trained weights, and trained with learning rate of 1e-4.

For \textbf{KITTI}, there are 7481 images in the annotated data. This data is divided intro train/test/val splits. 4500 images are used for training, 500 images are in validation set and rest 2481 are in the test dataset. 
For Temperature scaling(\cite{bosch-calib}), we use validation dataset to learn the temperature parameter. We train the temperature parameter until its value is stabilized. 

We have exactly same procedure for \textbf{Cityscapes} dataset as well. In Cityscapes, we have 3475 annotated images, out of which 2500 are used for training, 475 are used for validation, and 500 are used for testing the models. We perform similar holdout cross validation for Cityscapes also, and choose the best performing model for  testing. We use same procedure as \textbf{KITTI} for Temperature scaling.

\begin{table*}[tb]
\begin{adjustbox}{width=0.95\textwidth}

\begin{tabular}{c|c|c|c|c|c}
\toprule
\textbf{Approach}                              & \textbf{SmoothL1 (GT)} & \textbf{SmoothL1} & \textbf{ECE(z)} & \textbf{ECE(q)} & \textbf{NLL} \\
\midrule

\textbf{NLL Loss}\cite{loss_att}                 & $1.67 \pm 0.35$ & 	$2.16 \pm 0.36$ & 	$0.016 \pm 0.004$ & 	$0.854 \pm 0.109$ & 	$-1.31 \pm 0.14$   \\
\textbf{Calibration Loss}\cite{bosch-calib}                 & $1.68 \pm 0.32$ & 	$2.19 \pm 0.31$ & 	$0.013 \pm 0.004$ & 	$0.72 \pm 0.232$ & 	$-1.36 \pm 0.20$   \\
\textbf{Temperature Scaling}\cite{bosch-calib}              & $1.67 \pm 0.35$ & 	$2.16 \pm 0.36$ & 	$0.007 \pm 0.001$ & 	$0.128 \pm 0.098$ & 	$-1.38 \pm 0.12$    \\
\textbf{Isotonic Regression}\cite{isotonic}	     & $1.60 \pm 0.32$   &  	$2.05 \pm 0.31$   &  	$0.019 \pm 0.005$   &  	$0.909 \pm 0.049$   &  	$-1.31 \pm 0.16$ \\
\textbf{GP-Beta}\cite{gp-beta}                          & $1.60 \pm 0.31$ & 	$2.06 \pm 0.31$ & 	$0.018 \pm 0.009$ & 	$0.837 \pm 0.147$ & 	$-1.31 \pm 0.20$ \\
\textbf{$f$-Cal-KL (ours)}                              & $1.67 \pm 0.32$ & 	$2.16 \pm 0.32$ & 	$0.005 \pm 0.001$ & 	$0.068 \pm 0.007$ & 	$-1.43 \pm 0.08$   \\
\textbf{$f$-Cal-Wass (ours)}                            & $1.59 \pm 0.28$ & 	$2.08 \pm 0.28$ & 	$0.006 \pm 0.001$ & 	$0.037 \pm 0.0022$ & 	$-1.45 \pm 0.05$ \\

\midrule

\textbf{NLL Loss}\cite{loss_att}	     &   $1.44 \pm 0.34$   &  	$1.54 \pm 0.34$   &  	$0.0173 \pm 0.0027$   &  	$0.9183 \pm 0.0238$   &  	$-1.60 \pm 0.21$ \\
\textbf{Calibration Loss}\cite{bosch-calib}	     &   $1.46 \pm 0.29$   &  	$1.57 \pm 0.31$   &  	$0.0113 \pm 0.0022$   &  	$0.7611 \pm 0.0129$   &  	$-1.68 \pm 0.19$ \\
\textbf{Temperature Scaling}\cite{bosch-calib}	     & $1.44 \pm 0.34$   &  	$1.54 \pm 0.34$   &  	$0.0082 \pm 0.0019$   &  	$0.0922 \pm 0.0235$   &  	$-1.70 \pm 0.15$ \\
\textbf{Isotonic Regression}\cite{isotonic}	     & $1.38 \pm 0.27$   &  	$1.49 \pm 0.27$   &  	$0.0205 \pm 0.0045$   &  	$0.9378 \pm 0.0124$   &  	$-1.57 \pm 0.24$ \\
\textbf{GP-Beta}\cite{gp-beta}                    &   $1.39 \pm 0.26$   &  	$1.49 \pm 0.27$  &  	$0.0221 \pm 0.0082$  &  	$0.9348 \pm 0.0208$  &  	$-1.54 \pm 0.28$\\
\textbf{$f$-Cal-KL (ours)}                      &    $1.42 \pm 0.33$    & 	$1.52 \pm 0.33$   & 	$0.0056 \pm 0.0011$   & 	$0.0921 \pm 0.0135$   & 	$-1.76 \pm 0.15$ \\
\textbf{$f$-Cal-Wass (ours)}                      &    $1.43 \pm 0.34$ & 	$1.54 \pm 0.34$ & 	$0.0079 \pm 0.0016$ & 	$0.0799 \pm 0.0095$ & 	$-1.75 \pm 0.15$  \\

\bottomrule
\end{tabular}
\end{adjustbox}
\vspace{0.5em}
\caption{\textbf{Bokeh - disc-tracking}: We evaluate several baselines under homoscedastic (top-half) and heteroscedastic noise (bottom-half). \coolname{} is consistently better calibrated (lower ECE) compared to all considered baselines outperforms all considered baselines. Notably, this improved calibration comes without any  in terms of calibration, without sacrificing regression performance. SmoothL1 and SmoothL1 (GT) scores have been scaled by $1000$ and ECE scores by 100. }
\label{table:toy_exp_full}
\end{table*}

\section{Additional Results}

\subsection{Bokeh }
\label{suppl:bokeh_results}
\begin{figure*}
\begin{center}
    \centering
    \includegraphics[width=0.8\textwidth]{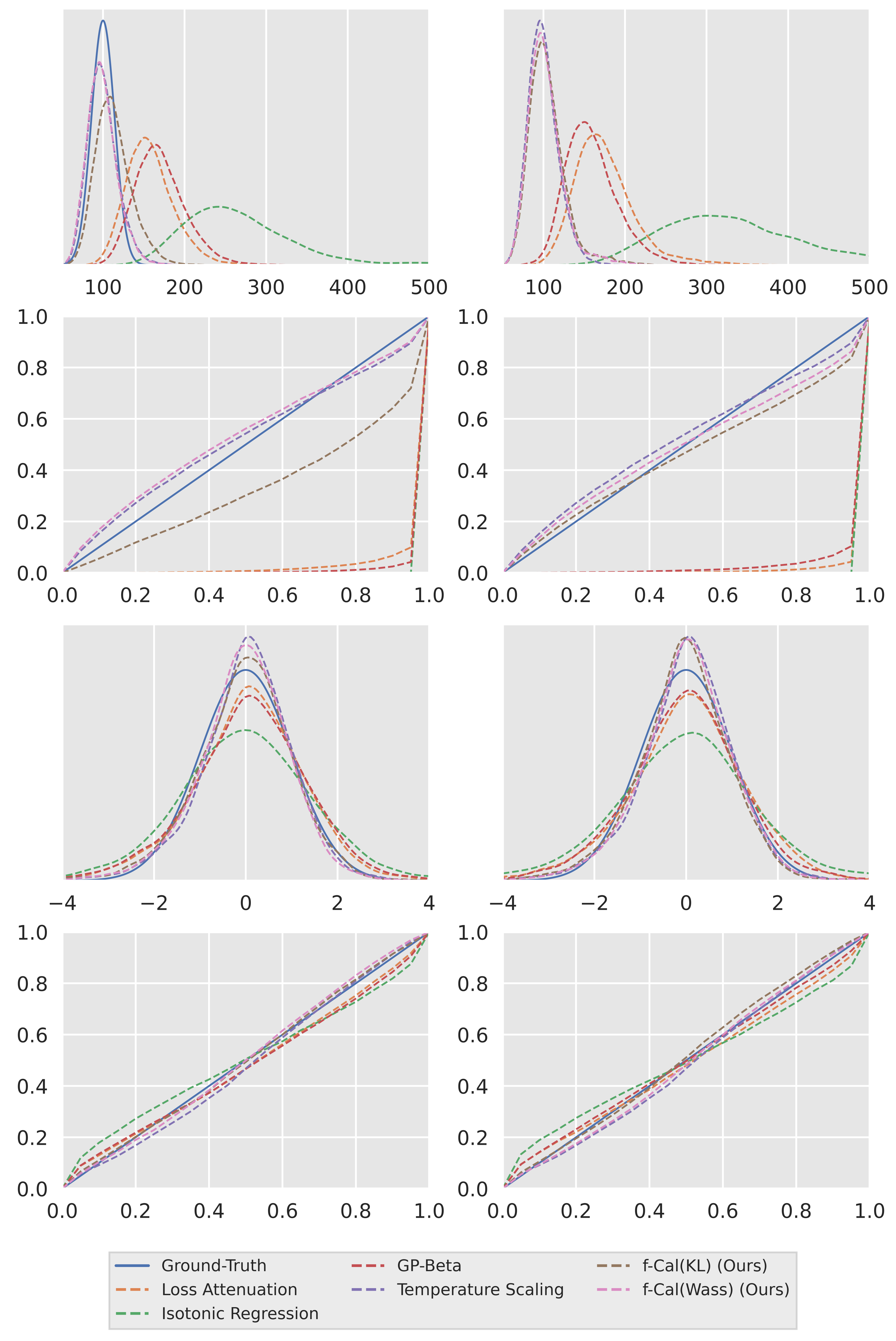}
    \caption{\textbf{Distributional and Reliability Diagrams (Bokeh)}: (Col 1) with Homoscedastic Noise (Col 2) with Heteroscedastic noise. (Row 1) shows the chi-squared distributional comparison of the predicted outputs with the target. (Row 2) shows the reliability diagram with chi-squared distribution. (Row 3) shows the standard-normal distributional comparison. (Row 4) shows the reliability curve with standard-normal variables.}
    \label{fig:toy_reliability}
\end{center}
\end{figure*}

Table \ref{table:toy_exp_full} shows the results for \coolname{} and the baseline calibration techniques on Bokeh dataset with both homoscedastic and heteroscedastic noise. It can be seen that \coolname{} outperforms all the baseline methods on all calibration metrics while maintaining the similar or sometimes better deterministic performance as the base model(i.e NLL loss). From Fig. \ref{fig:toy_reliability}, we can see qualitatively that the output distributions from \coolname{} trained models are much closer to the ground truth distribution than the baselines.  We can also see from the reliability curves, that \coolname{} models are much closer to the diagonal line representing perfect calibration than the baselines. Apart from the baseline methods shown in Table\ref{table:toy_exp_full}, we also trained MMD\cite{cui2020calibrated}, but the model fails to converge on Bokeh which is the simplest of the three datasets used in this paper. 

\subsection{Object detection}
\label{suppl:object_detection}

In this section, we extensively report qualitative results on Object detection with \coolname{}. In the main paper, we reported results for Expected calibration error(ECE) and Negative log likelihood(NLL). Here, we report results on other metrics such as Maximum calibration error, KLD, Wasserstein distance between proposed and target distributions.  

The above metrics reflect calibration quality of the models. In addition, for object detection, we also modify existing popular metrics such as mAP(\cite{lin2014microsoft}) and PDQ(\cite{hall2020probabilistic}) to report consistency of the models. Calibration generally implies consistency(vice versa is not true). In the main paper, we have defined calibration. Here we define consistency as below,

\begin{definition}[\textbf{Consistency:}]
A neural regressor $f_p$ is consistent for any arbitrary confidence bound $c$ if,
\begin{equation}
    p(Y \leq y | s(y)) \leq c
\end{equation}
\label{definition:consistency}
\end{definition}

From the above definition, we can observe that the requirement for calibration is more stringent than that of consistency. We can have consistent probabilistic detection if we predict arbitrarily high uncertainty(Intro figure(c)), we can always have consistent predictions, though the uncertainties can not be interpreted as confidence scores. Through these new metrics, we show that consistency is the byproduct of calibration. We do not enforce any explicit constraints for consistency, yet we show in our results that we end up achieving highly consistent prediction. It is important to note that the consistency metrics we report do not automatically imply calibration. So these metrics should be interpreted in conjunction with calibration metrics. We can have arbitrarily high consistency if we predict highly inflated uncertainty estimates.

Towards this end, we modify two popular object detection evaluation metrics, mAP and PDQ, for consistency estimation. The new metrics are minor modification of mAP and PDQ, designed to evaluate consistency of the object detector. We use definition \ref{definition:consistency} to build these metrics. 

Formally, let's say our groundtruth bounding box is $B_g = [x_1^{g}, y_1^{g}, x_2^{g}, y_2^{g}]$, represented by top left and bottom right corners of the bounding box. Our predicted box is represented by $B_d = \mathcal{N}(\mu_d, \Sigma_d)$. Here, $\mu_d = [\mu_{x_1}^{d}, \mu_{y_1}^{d}, \mu_{x_2}^{d}, \mu_{y_2}^{d}]$. $\Sigma_d$ is 4 x 4 matrix representing co-variance matrix for bounding box. It can be a full co-variance of diagonal covariance too. For this work, we are assuming diagonal co-variance however our proposed metric will be applicable to full co-variance matrix. Given this, the loss attenuation formulation looks as below,

\begin{equation}
    \label{loss_att}
    L_{la} = (B_g - \mu_d)^T \Sigma_d^{-1} (B_g - \mu_d) + \log(det(\Sigma_d)) \\
\end{equation}

In equation \ref{loss_att}, first term represents squared Mahalanobis distance, which charaterizes number of standard deviations a point is away from
mean of a distribution. The squared mahalanobis distance followed chi-squared distribution with p degress of freedom. We get mahalanobis distance threshold $\mathcal{M}_{thresh}$ when we evaluate chi-squared distribution with p degrees of freedom and confidence interval $\alpha$. In this case, it will denote the probability of the groundtruth being in the hyper-ellipse defined by squared Mahalanobis distance. 

In this work, we propose to use probability confidence between groundtruth and predicted distribution as a quality measure for detection. The less the mahalanobis distance, the more close ground-truth and the distribution are, the smaller the confidence contour would be. 

Now we formally define confidence contour and corresponding Mahalanobis distance. Let's say confidence contour of a Gaussian distribution's volume of $\alpha$. It means that probability of a random variable X falling inside this confidence contour is $\alpha$. In this case, probability of exceeding critical value is $\beta = 1 - \alpha$. Then Squared Mahalanobis distance threshold is $\mathcal{M}_{thresh} = \tilde{\chi}^2(p, \beta)$. Which means that for a normal distribution $\mathcal{N}(\mu, \Sigma)$, if X falls in confidence contour $\alpha$, then $(X- \mu)^T\Sigma^{-1}(X - \mu) \leq \tilde{\chi}^2(p, 1 - \alpha)$. \\

\begin{figure*}[h]
\begin{center}
    \centering
    \includegraphics[width=1.0\columnwidth]{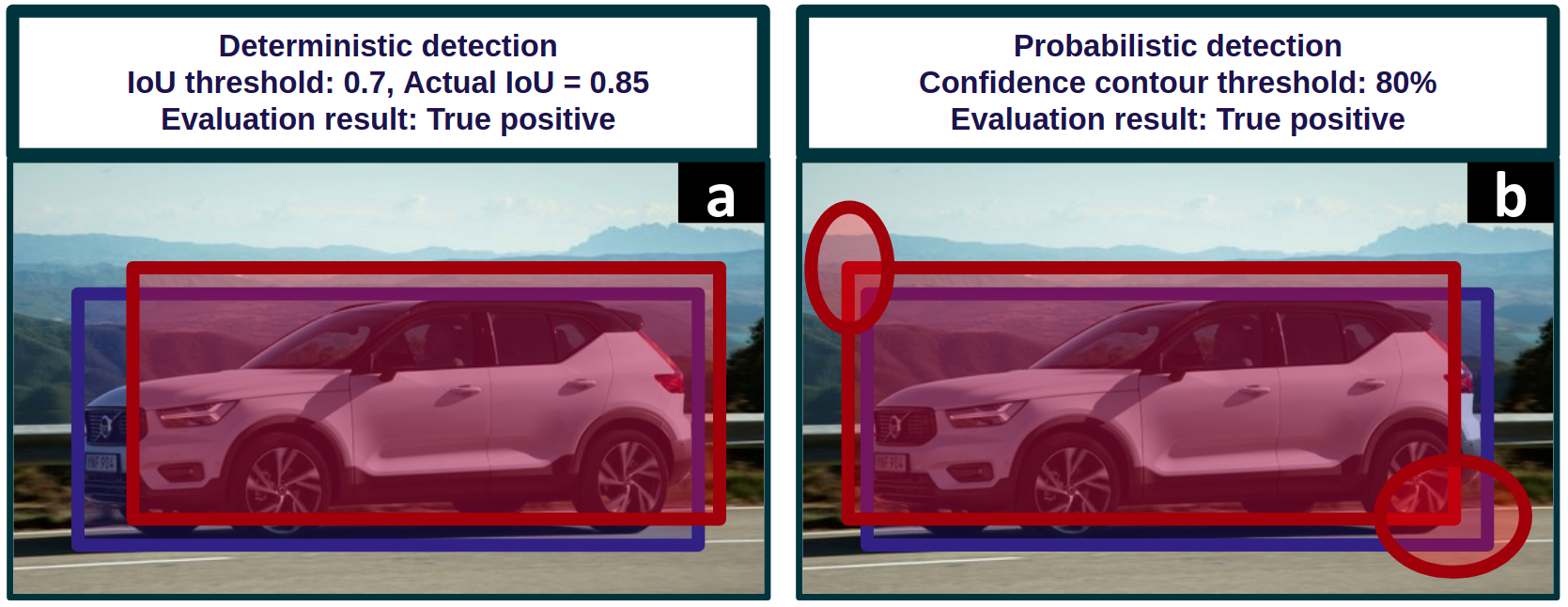}
    \vspace{0.3em}
    \caption{The definition of True positive changes as we define confidence contour based criterion. Blue box represents groundtruth object in both the images. In deterministic detection(a), we evaluate certain proposal as true positive if the IoU with actual object is greater than certain threshold. In probabilistic detection(b), we represent the uncertainty with ellipses in the figure. The ellipses visualized correspond to 80\% confidence contour, and we observe that the groundtruth falls within 80\% confidence contour, hence we classify that probabilistic detection as a true positive.}
    \label{fig:prob_eval}
\end{center}
\end{figure*}

\noindent
\textbf{\textit{Mean Mahalanobis average precision(mMAP):}}

Mean average precision(mAP)\cite{lin2014microsoft} has been the most popular metric to evaluate object detector. For consistency estimation, we modify this metric to incorporate probabilistic bounding box predictions. In mAP, precision is calculated for IoUs of $0.5$ to $0.95$ at the interval of $0.05$. In mMAP, we replace IoU threshold with confidence contour thresholds. We keep confidence contour as a threshold and determine true positive based on Mahalanobis distance as explained in figure \ref{fig:prob_eval}. In this work, we have thresholds of [0.999, 0.995, 0.99 , 0.95, 0.9, 0.85, 0.8, 0.7]. We observe that mMAP is more informative metric to analyse probabilistic consistency. The definition of consistency(\ref{definition:consistency}) requires the prediction to be in certain confidence contour bound($c$). This metric just evaluates that over multiple thresholds and averages across thresholds and categories just like mAP.

\noindent
\textbf{\textit{Probability-based detection quality(PDQ):}}

\begin{figure*}
\begin{subfigure}{.48\textwidth}
  \centering
  \includegraphics[width=0.99\linewidth]{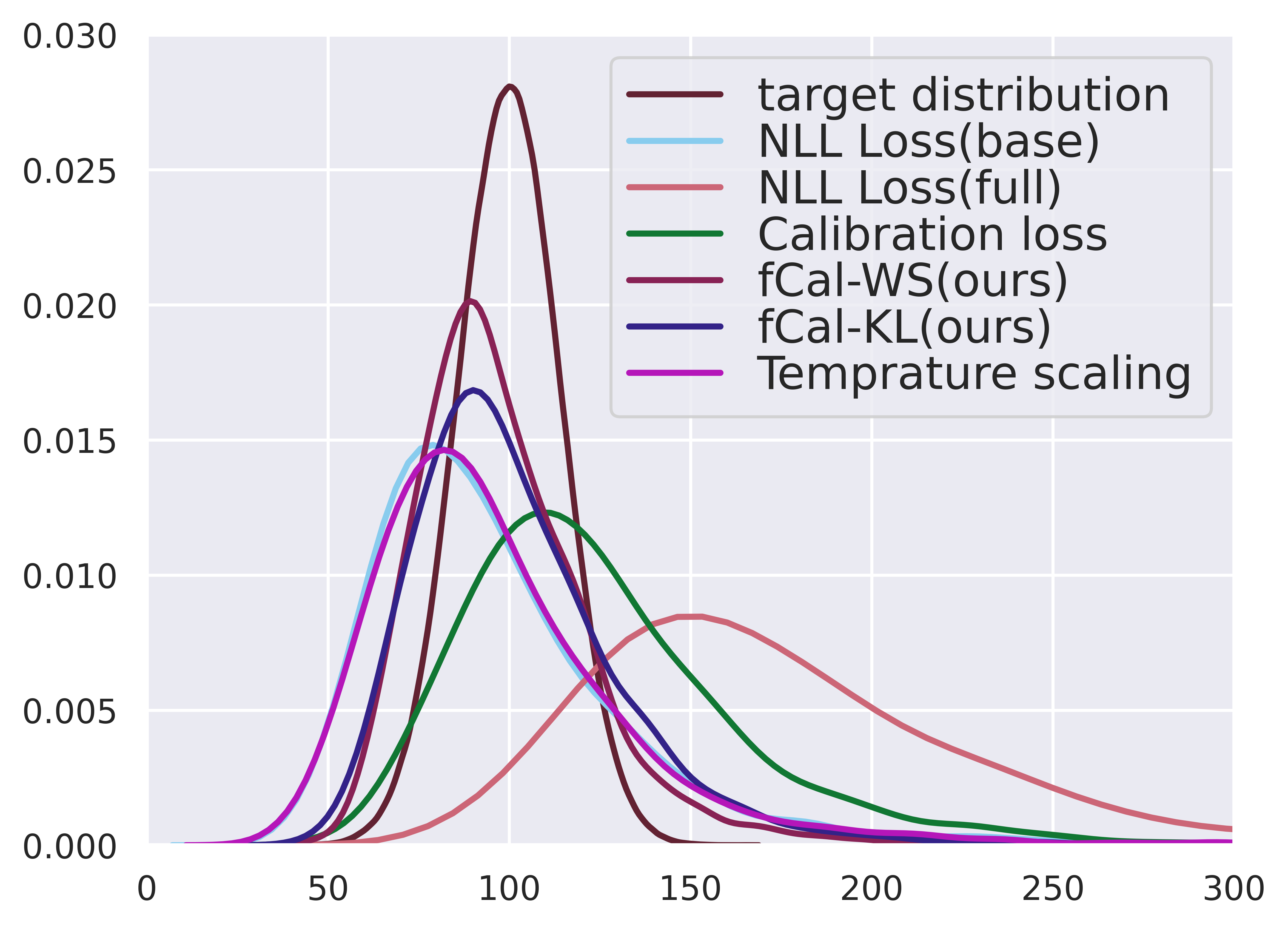}  
  \caption{Object detection - Cityscapes\cite{cordts2016cityscapes} - chisquared distribution plots of different baselines. To quantitatively understand the results, see \textbf{W-dist(q)} and \textbf{KLD(q)} rows of table \ref{table:cs-calibration}. Lower values correspond to better curves, which can be visually understood in the figure.}
  \label{fig:sub-first-cs}
\end{subfigure}
\begin{subfigure}{.48\textwidth}
  \centering
  \includegraphics[width=0.99\linewidth]{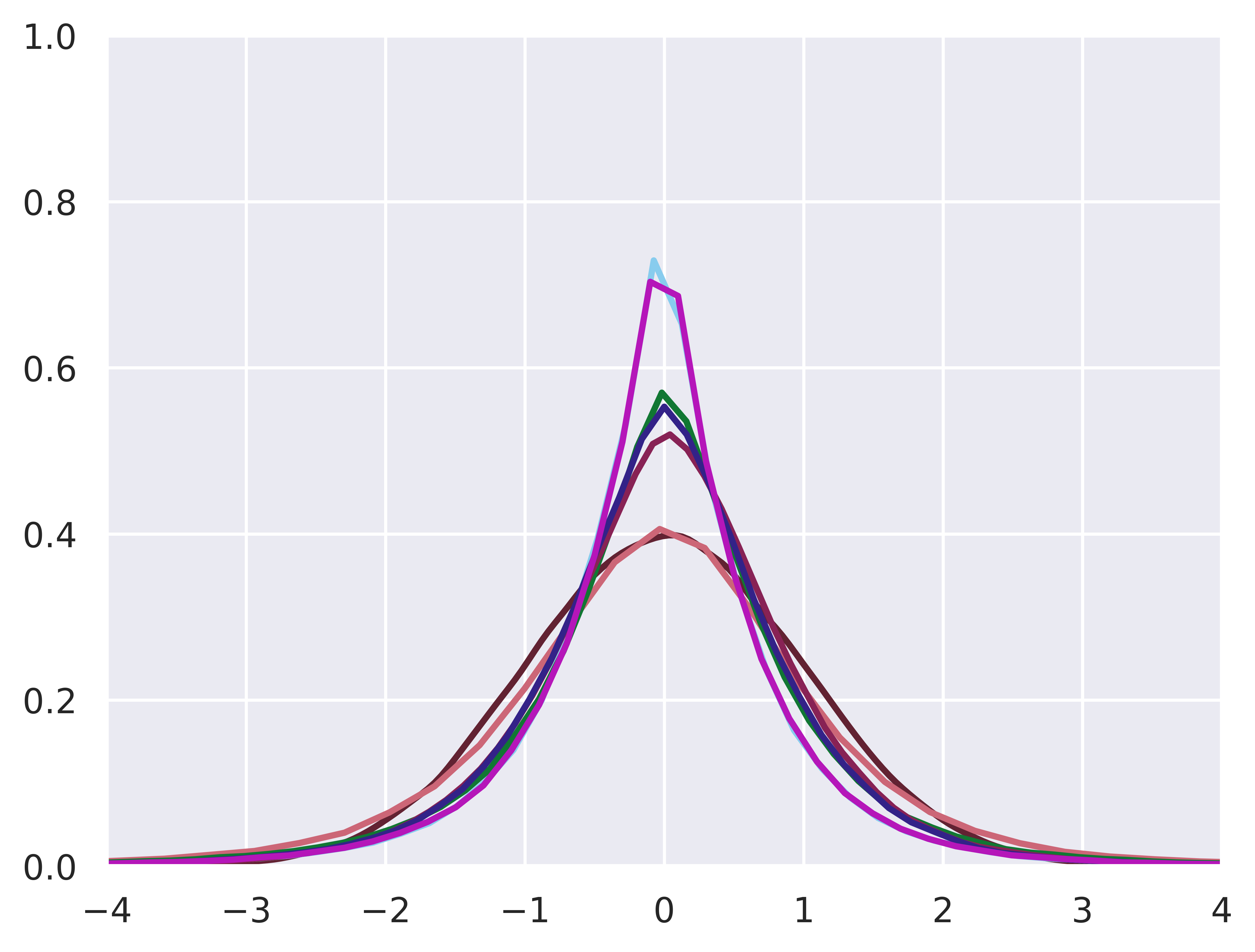}  
  \caption{Object detection - Cityscapes\cite{cordts2016cityscapes} - standard normal distribution plots of different baselines. To quantitatively understand the results, see \textbf{W-dist(z)} and \textbf{KLD(z)} rows of table \ref{table:cs-calibration}. }
  \label{fig:sub-second-cs}
\end{subfigure}
\newline
\begin{subfigure}{.48\textwidth}
  \centering
  \includegraphics[width=0.99\linewidth]{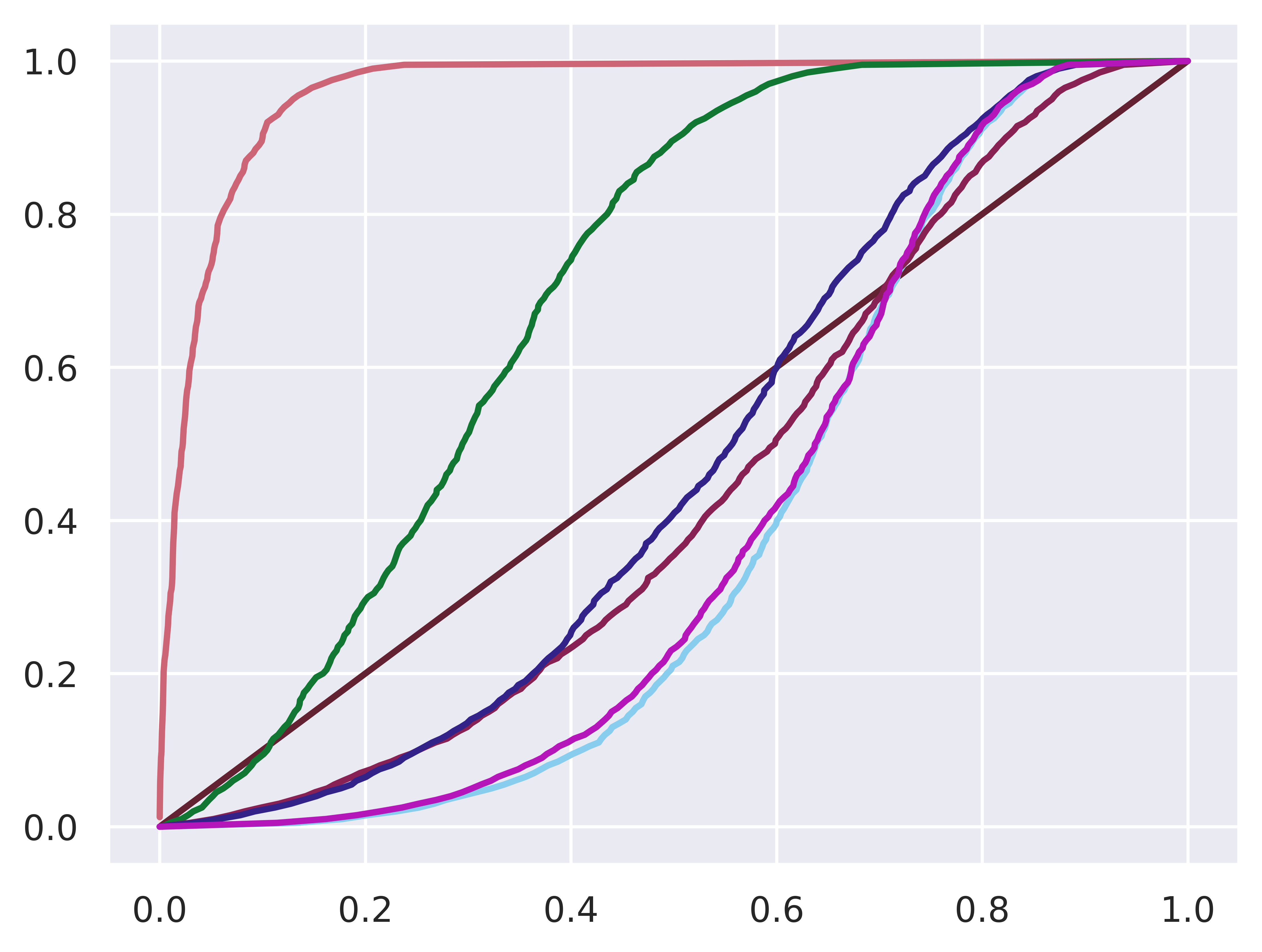}  
  \caption{Object detection - Cityscapes\cite{cordts2016cityscapes} - chisquared reliability diagrams of different baselines. To quantitatively understand the results, see \textbf{MCE(q)} and \textbf{ECE(q)} rows of table.  \ref{table:cs-calibration}. Lower values reflect better curves.}
  \label{fig:sub-third-cs}
\end{subfigure}
\begin{subfigure}{.48\textwidth}
  \centering
  \includegraphics[width=0.99\linewidth]{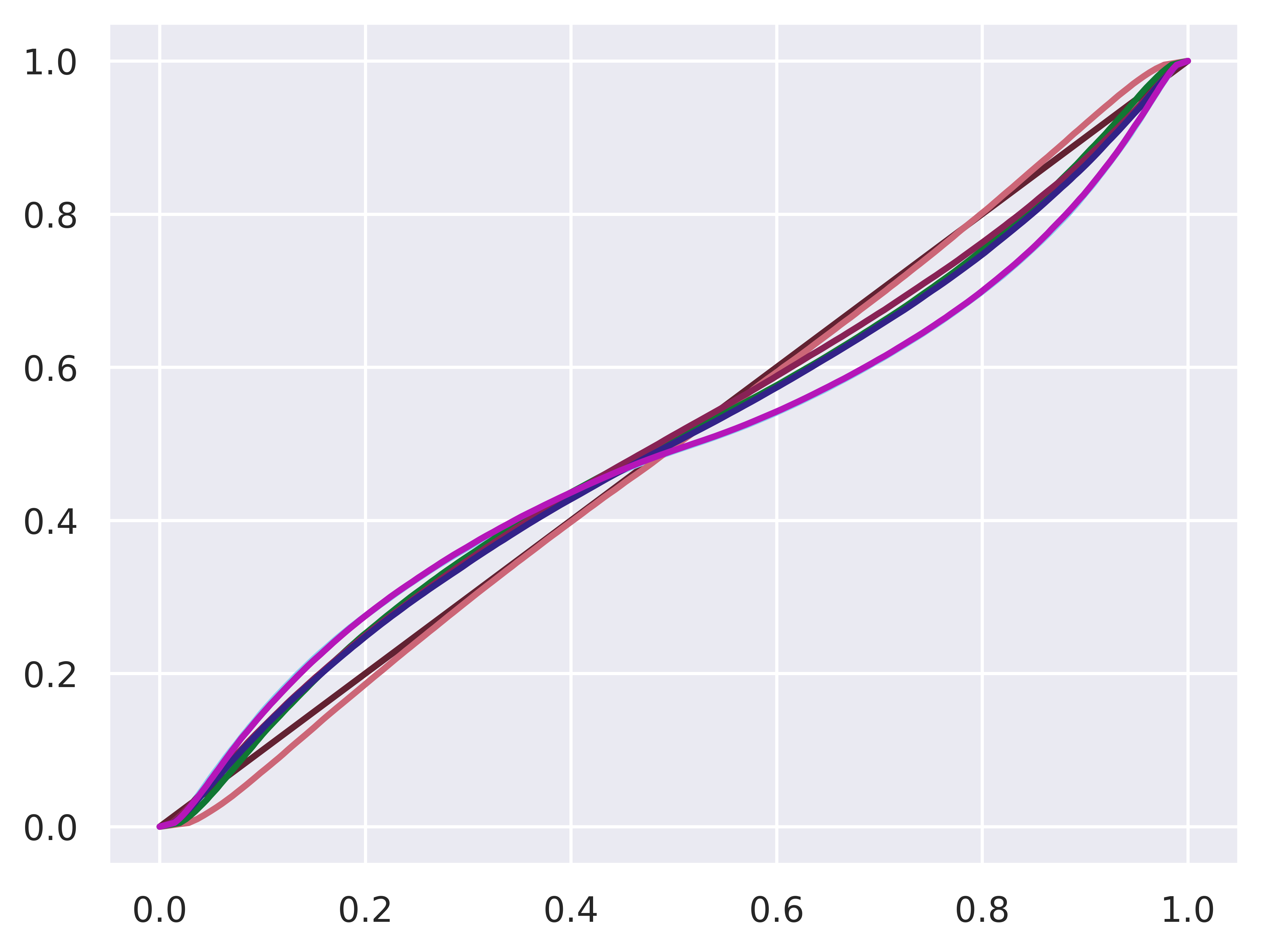}  
  \caption{Object detection - Cityscapes\cite{cordts2016cityscapes} - standard normal reliability diagrams of different baselines. To quantitatively understand the results, see \textbf{MCE(z)} and \textbf{ECE(z)} rows of table.  \ref{table:cs-calibration}. }
  \label{fig:sub-fourth-cs}
\end{subfigure}
\vspace{0.5em}
\caption{Reliability diagrams and distribution plots for Object detection - Cityscapes\cite{cordts2016cityscapes} dataset. }
\label{fig:cityscapes-all}
\end{figure*}

PDQ(Probability-based Detection Quality)\cite{hall2020probabilistic} is a recently proposed metric to evaluate Probabilistic object detectors. It uses spatial and label quality into the evaluation criteria, and explicitly rewards probabilistically accurate detections. Both Spatial and label quality measures are calculated between all possible pairs of detections and Groundtruth. Geometric mean of these two measures is calculated and used to find the optimal assignment between all detections and groundtruths. 

In PDQ, spatial quality is calculated by fusing background and foreground loss, which are computed using groundtruth segmentation mask and the probabilistic detection. This requires availability of masks during test time which may not be possible for bounding box based object detection dataset. In addition to that, evaluation objective of spatial quality estimation is different compared to training objective of probabilistic object detectors, which are trained with NLL loss. In contrast, the modified evaluation criteria(\ref{fig:prob_eval}) evaluates spatial quality without the need of any segmentation mask. Incorporating Mahalanobis distance based criteria enables the modified metric to evaluate consistency. Mahalanobis distance is a reflection of how many standard deviations away your mean prediction is compared to the sample(ground-truth in this case). Less Mahalanobis distance could be interpreted as good spatial quality of the prediction. We redefine the spatial quality as below,

\begin{equation}
\label{spatialquality}
    Q_s(B_g, B_d) = \exp( \frac{-(B_g - \mu_d)^T \Sigma_d^{-1} (B_g - \mu_d)}{T}) 
\end{equation}

Where $T$ is the temperature parameter(Not to be confused with Temperature parameter of Temperature scaling method). It determines how much should the Mahalanobis distance penalize spatial quality. Higher the temperature value, lower the penalty. In our experiments, we keep the value of $T$ to be 10. 

Both these metrics, complement mAP, which gives us estimate of how accurate our bounding boxes are, while these metrics will tell us how consistent our uncertainty values are. We can infer more about consistency of our models by analysing these metrics.

\begin{figure*}
\begin{subfigure}{.48\textwidth}
  \centering
  \includegraphics[width=0.99\linewidth]{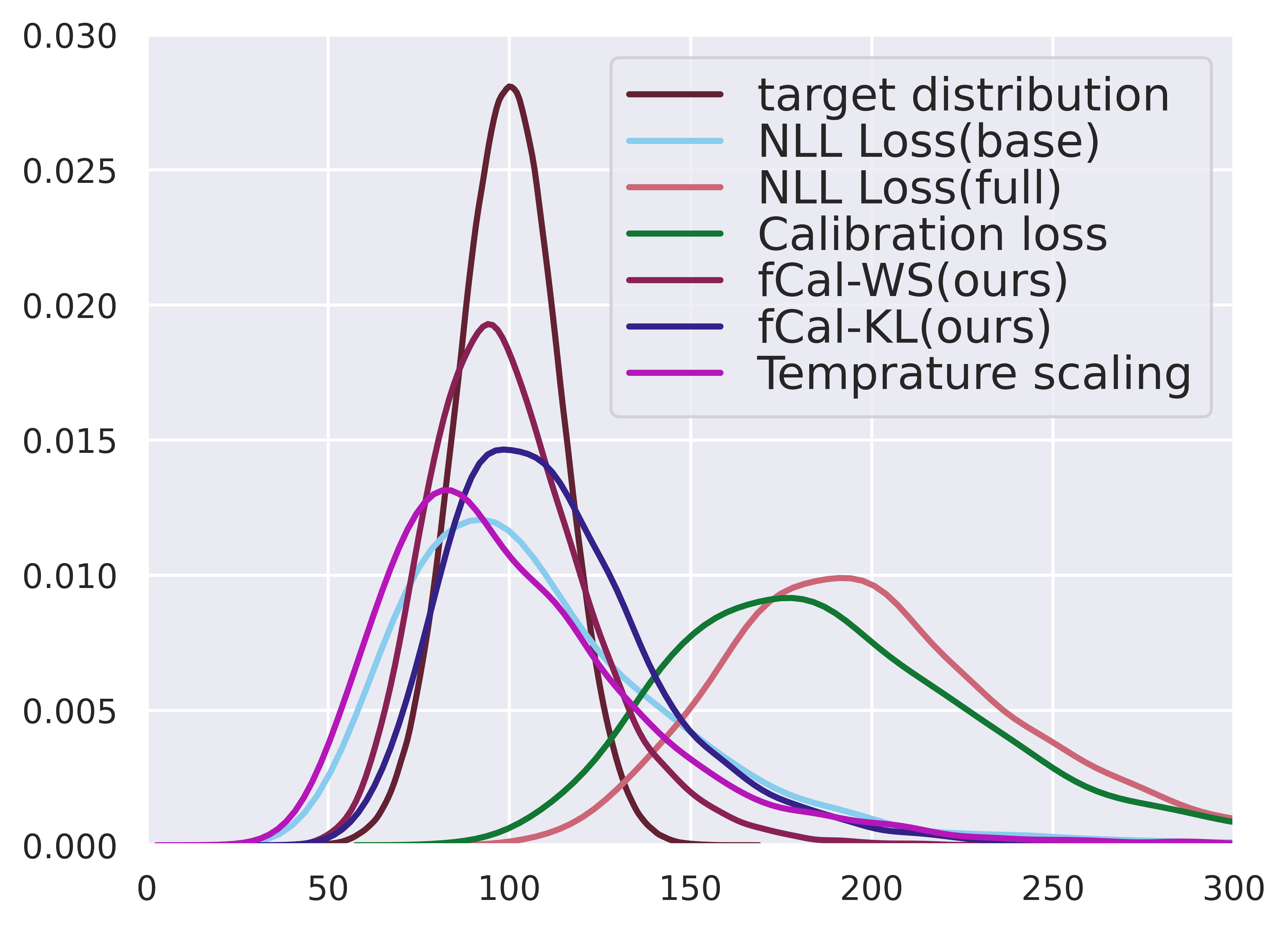}  
  \caption{Object detection - Kitti\cite{kitti} - chisquared distribution plots of different baselines. To quantitatively understand the results, see \textbf{W-dist(q)} and \textbf{KLD(q)} rows of table \ref{table:kitti-calibration}. Lower values correspond to better curves, which can be visually understood in the figure.}
  \label{fig:sub-first-kitti}
\end{subfigure}
\begin{subfigure}{.48\textwidth}
  \centering
  \includegraphics[width=0.99\linewidth]{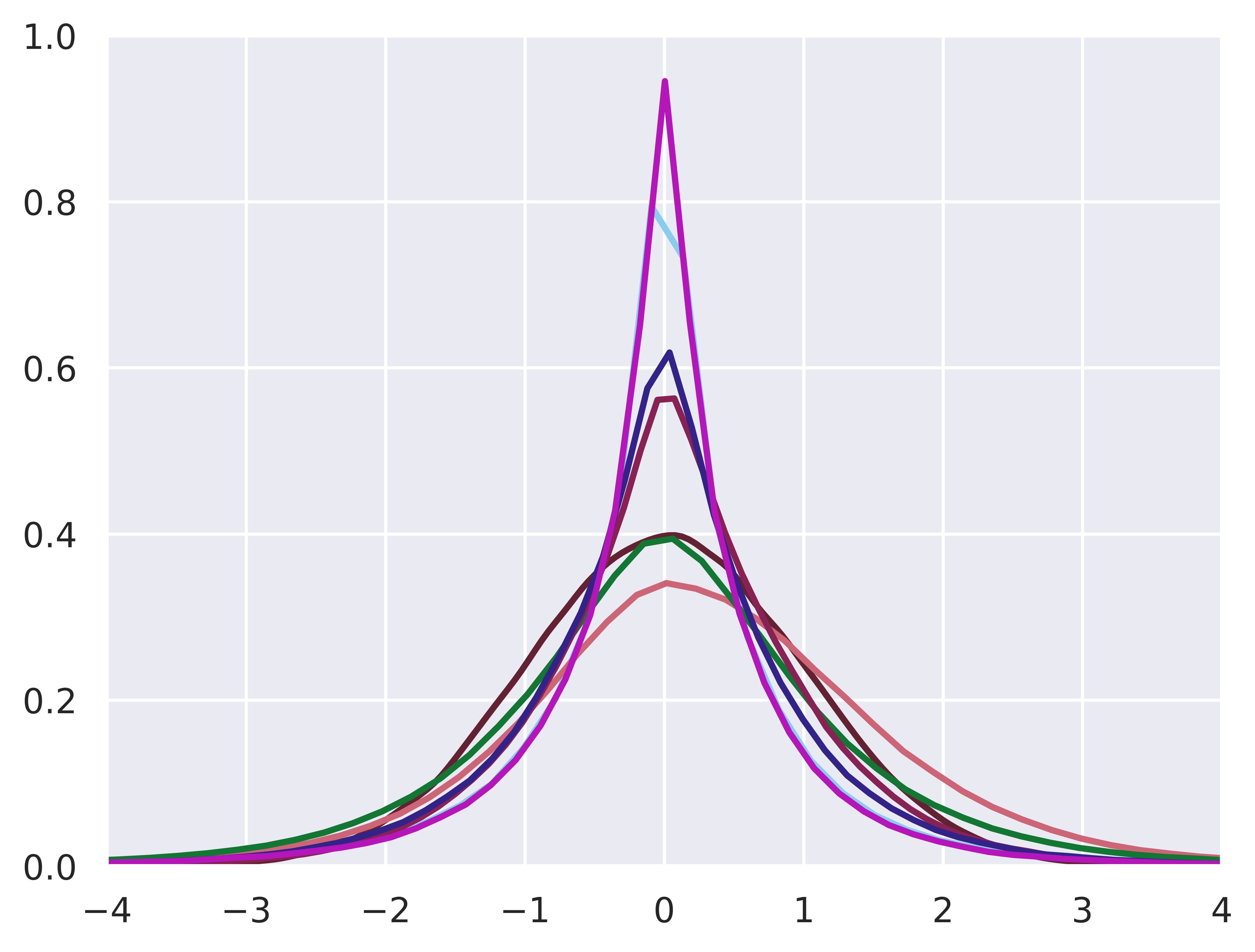}  
  \caption{Object detection - Kitti\cite{kitti} - standard normal distribution plots of different baselines. To quantitatively understand the results, see \textbf{W-dist(z)} and \textbf{KLD(z)} rows of table \ref{table:kitti-calibration}.}
  \label{fig:sub-second-kitti}
\end{subfigure}
\newline
\begin{subfigure}{.48\textwidth}
  \centering
  \includegraphics[width=0.99\linewidth]{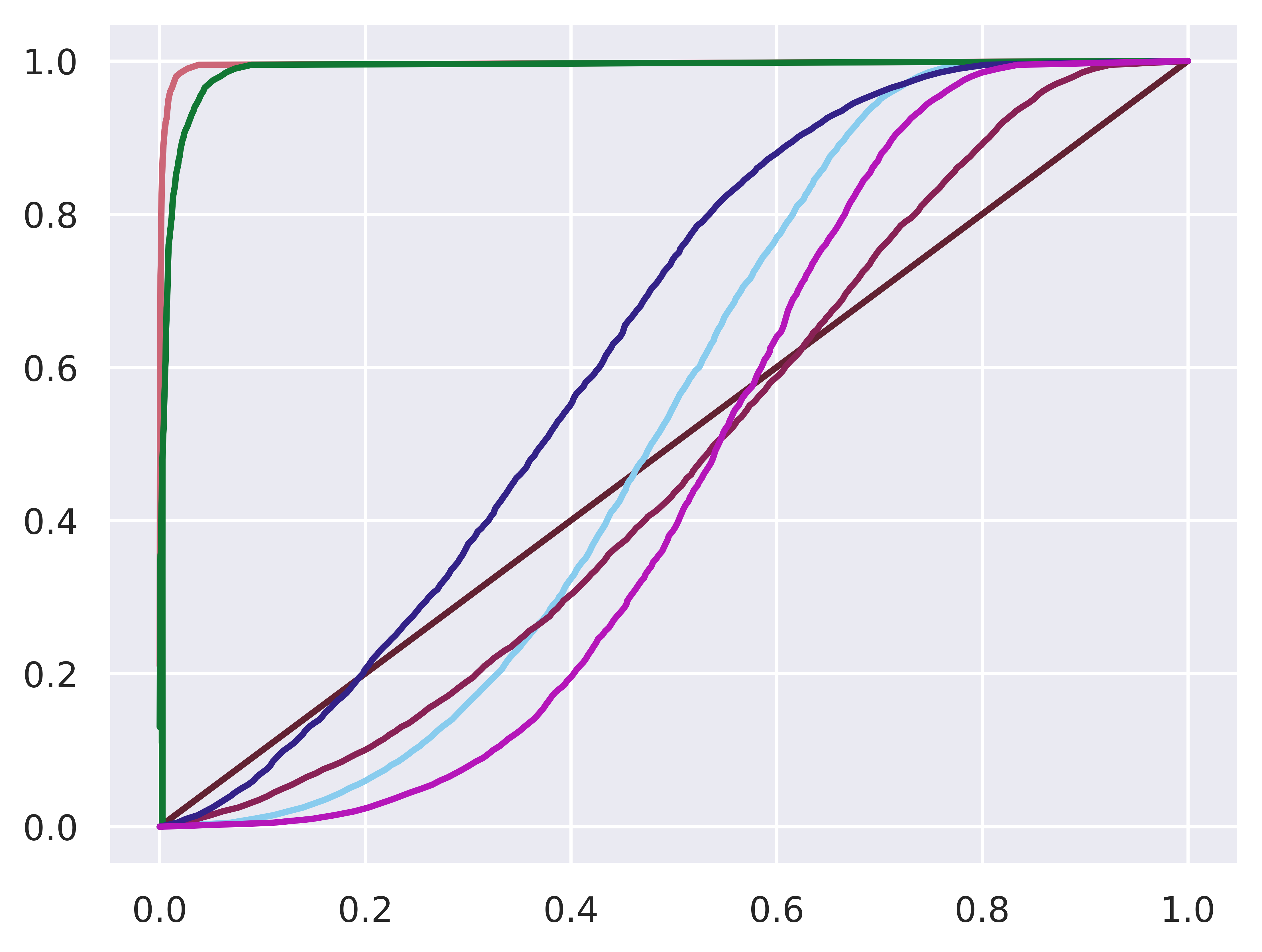}  
  \caption{Object detection - Kitti\cite{kitti} - chisquared reliability diagrams of different baselines. To quantitatively understand the results, see \textbf{MCE(q)} and \textbf{ECE(q)} rows of table \ref{table:kitti-calibration}. Lower values reflect better curves.}
  \label{fig:sub-third-kitti}
\end{subfigure}
\begin{subfigure}{.48\textwidth}
  \centering
  \includegraphics[width=0.99\linewidth]{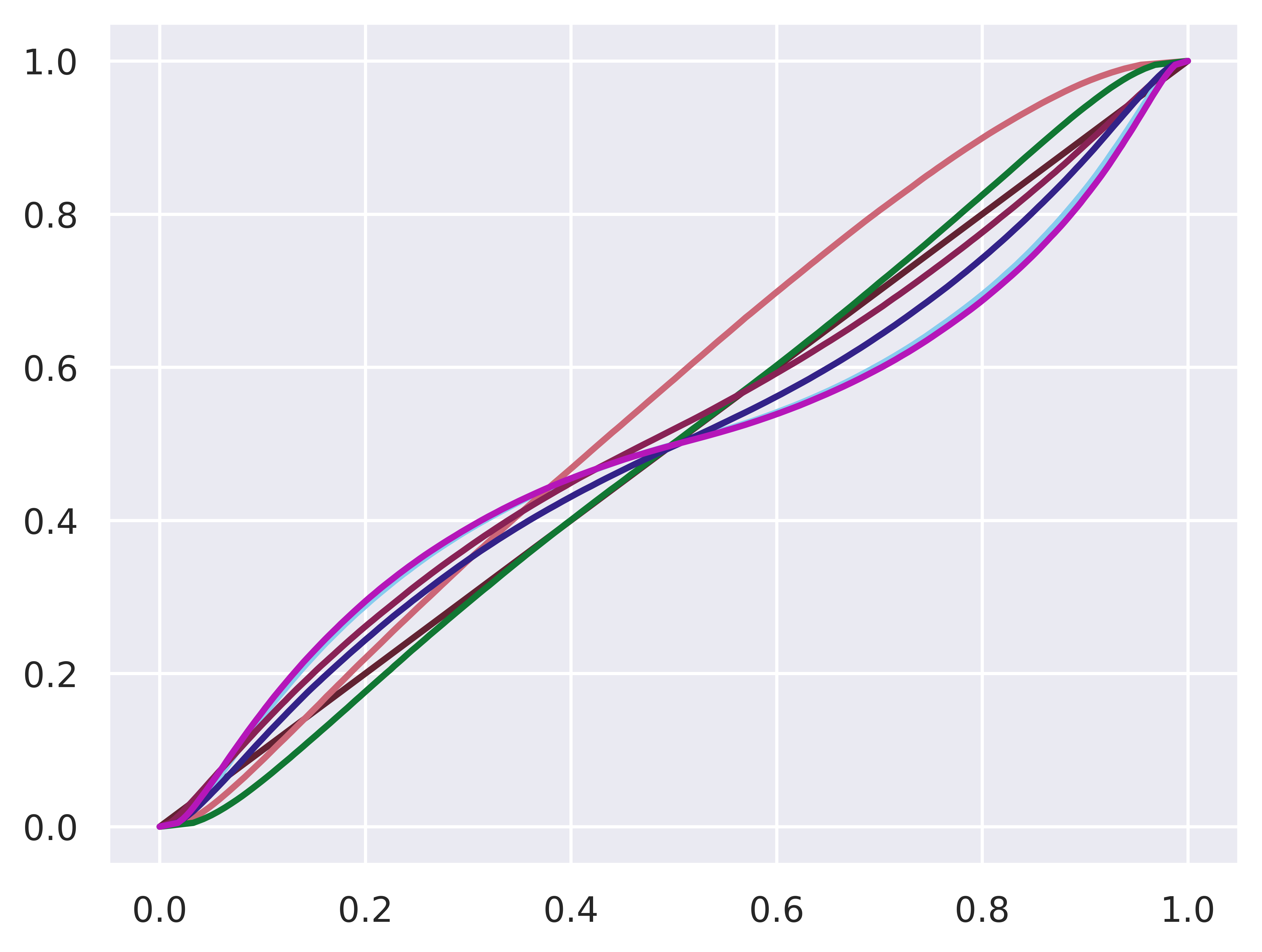}  
  \caption{Object detection - Kitti\cite{kitti} - standard normal reliability diagrams of different baselines. To quantitatively understand the results, see \textbf{MCE(z)} and \textbf{ECE(z)} rows of table \ref{table:kitti-calibration}.}
  \label{fig:sub-fourth-kitti}
\end{subfigure}
\vspace{0.5em}
\caption{Reliability diagrams and distribution plots for Object detection - Kitti\cite{kitti} dataset. }
\label{fig:kitti-all}
\end{figure*}

\noindent
\textbf{\textit{Cityscapes results:}}

In tables \ref{table:cs-consistency} and \ref{table:cs-calibration}, we report results for Cityscapes\cite{cordts2016cityscapes} dataset. We observe that \coolname{} is able to obtain highly consistent and calibrated uncertainty estimates. We observe that NLL Loss(full) is providing overconfident uncertainty estimates, resulting in poor consistency. We observe that other baselines such as NLL Loss(base) and Calibration Loss are able to obtain high consistency, but poorer calibration, which is a result of inflated uncertainty estimates. \coolname{} and Temperature scaling are able to yield calibrated and consistent uncertainty estimates, while retaining deterministic performance. However, Temperature scaling had holdout validation set for tuning temperature parameter, while \coolname{} results are directly obtained using training data only. We also note that \coolname{} enables the model to learn uncertainty aware representations, as opposed to Temperature scaling, where representations learned are same as those of uncalibrated models.

\noindent
\textbf{\textit{KITTI results:}}

\begin{table}[]
\centering
\resizebox{\textwidth}{!}{%
\begin{tabular}{|c|c|c|c|c|c|c|c|c|}
\hline
\textbf{Approach} &
  \textbf{mAP} &
  \textbf{AP50} &
  \textbf{AP75} &
  \textbf{mMAP} &
  \textbf{PDQ} &
  \textbf{\begin{tabular}[c]{@{}c@{}}PDQ\\ (spatial)\end{tabular}} &
  \textbf{\begin{tabular}[c]{@{}c@{}}PDQ\\ (label)\end{tabular}} &
  \textbf{NLL} \\ \hline
\textbf{\begin{tabular}[c]{@{}c@{}}NLL Loss\\ (base)\end{tabular} \cite{loss_att} }     & 54.451 & 78.476 & 62.876 & 76.76  & 0.601 & 0.725 & 0.976 & 1.022 \\ \hline
\textbf{\begin{tabular}[c]{@{}c@{}}NLL Loss\\ (full)\end{tabular} \cite{loss_att}}     & 51.764 & 76.245 & 58.893 & 35.066 & 0.443 & 0.483 & 0.975 & 0.932 \\ \hline
\textbf{\begin{tabular}[c]{@{}c@{}}Calibration \\ loss\end{tabular} \cite{bosch-calib}}   & 50.404 & 70.442 & 57.963 & 49.162 & 0.449 & 0.511 & 0.968 & 0.773 \\ \hline
\textbf{\begin{tabular}[c]{@{}c@{}}fCal-WS\\ (ours)\end{tabular}}      & 48.04  & 77.768 & 53.107 & 73.525 & 0.565 & 0.703 & 0.968 & 0.914 \\ \hline
\textbf{\begin{tabular}[c]{@{}c@{}}fCal-KL\\ (ours)\end{tabular}}      & 51.874 & 76.377 & 59.181 & 70.503 & 0.535 & 0.665 & 0.96  & 0.846 \\ \hline
\textbf{\begin{tabular}[c]{@{}c@{}}Temperature \\ scaling\end{tabular} \cite{bosch-calib}} & 54.451 & 78.476 & 62.876 & 77.433 & 0.608 & 0.737 & 0.976 & 1.021 \\ \hline
\end{tabular} }
\vspace{0.5em}
\caption{Object detection - KITTI \cite{kitti}: Consistency and deterministic results. This table shows results against various evaluation metrics for consistency and deterministic performance.}
\label{table:kitti-consistency}
\end{table}

\begin{table}[]
\centering
\resizebox{\textwidth}{!}{%
\begin{tabular}{|c|c|c|c|c|c|c|c|c|}
\hline
\textbf{Approach} & \textbf{ECE(z)} & \textbf{MCE(z)} & \textbf{ECE(q)} & \textbf{MCE(q)} & \textbf{W-dist(z)} & \textbf{KLD(z)} & \textbf{W-dist(q)} & \textbf{KLD(q)} \\ \hline
\textbf{\begin{tabular}[c]{@{}c@{}}NLL Loss\\ (base)\end{tabular} \cite{loss_att}}     & 0.00304 & 0.01396 & 0.0537  & 0.21358 & 0.005 & 0.004 & 1183.549  & 0.768 \\ \hline
\textbf{\begin{tabular}[c]{@{}c@{}}NLL Loss\\ (full)\end{tabular} \cite{loss_att} }     & 0.00345 & 0.0406  & 0.93312 & 0.96343 & 0.227 & 0.11  & 12190.394 & 3.052 \\ \hline
\textbf{\begin{tabular}[c]{@{}c@{}}Calibration \\ loss\end{tabular} \cite{bosch-calib}}   & 0.00233 & 0.02759 & 0.8261  & 0.90619 & 0.151 & 0.088 & 10065.516 & 2.435 \\ \hline
\textbf{\begin{tabular}[c]{@{}c@{}}fCal-WS\\ (ours)\end{tabular}}      & 0.00115 & 0.00697 & 0.00697 & 0.06596 & 0.003 & 0.002 & 78.076    & 0.174 \\ \hline
\textbf{\begin{tabular}[c]{@{}c@{}}fCal-KL\\ (ours)\end{tabular}}      & 0.00162 & 0.01175 & 0.0393  & 0.18911 & 0.005 & 0.004 & 528.606   & 0.517 \\ \hline
\textbf{\begin{tabular}[c]{@{}c@{}}Temperature \\ scaling\end{tabular} \cite{bosch-calib}} & 0.00315 & 0.01268 & 0.04126 & 0.16199 & 0.002 & 0.001 & 742.99    & 0.631 \\ \hline
\end{tabular} }
\vspace{0.3em}
\caption{Object detection - KITTI \cite{kitti}: Results of \coolname{} and other baselines for various calibration metrics. }
\label{table:kitti-calibration}
\end{table}

\begin{table}[h]
\vspace{0.5cm}
\centering
\resizebox{\textwidth}{!}{%
\begin{tabular}{|c|c|c|c|c|c|c|c|c|}
\hline
\textbf{method} &
  \textbf{mAP} &
  \textbf{AP50} &
  \textbf{AP75} &
  \textbf{mMAP} &
  \textbf{PDQ} &
  \textbf{\begin{tabular}[c]{@{}c@{}}PDQ\\ (spatial)\end{tabular}} &
  \textbf{\begin{tabular}[c]{@{}c@{}}PDQ\\ (label)\end{tabular}} &
  \textbf{NLL} \\ \hline
\textbf{\begin{tabular}[c]{@{}c@{}}NLL Loss\\ (base)\end{tabular} \cite{loss_att} }     & 38.309 & 61.548 & 39.142 & 49.380 & 0.454 & 0.613 & 0.910 & 1.069 \\ \hline
\textbf{\begin{tabular}[c]{@{}c@{}}NLL Loss\\ (full)\end{tabular} \cite{loss_att} }     & 36.199 & 55.878 & 39.283 & 23.763 & 0.361 & 0.453 & 0.934 & 1.029 \\ \hline
\textbf{\begin{tabular}[c]{@{}c@{}}Calibration\\  loss\end{tabular} \cite{bosch-calib} }   & 39.218 & 61.922 & 40.220 & 42.302 & 0.424 & 0.560 & 0.920 & 0.999 \\ \hline
\textbf{\begin{tabular}[c]{@{}c@{}}fCal-WS\\ (ours)\end{tabular}}      & 37.220 & 61.486 & 38.469 & 46.202 & 0.442 & 0.593 & 0.911 & 1.007 \\ \hline
\textbf{\begin{tabular}[c]{@{}c@{}}fCal-KL\\ (ours)\end{tabular}}      & 38.481 & 61.924 & 40.210 & 43.982 & 0.442 & 0.584 & 0.915 & 0.929 \\ \hline
\textbf{\begin{tabular}[c]{@{}c@{}}Temperature \\ scaling\end{tabular} \cite{bosch-calib} } & 38.309 & 61.548 & 39.142 & 48.965 & 0.452 & 0.610 & 0.911 & 1.065 \\ \hline
\end{tabular} }
\vspace{0.5em}
\caption{Object detection - Cityscapes\cite{cordts2016cityscapes}: Consistency and deterministic results. This table shows results against various evaluation metrics for consistency and deterministic performance.}
\label{table:cs-consistency}
\end{table}

\begin{table}[]
\centering
\resizebox{\textwidth}{!}{%
\begin{tabular}{|c|c|c|c|c|c|c|c|c|}
\hline
\textbf{Appoach} & \textbf{ECE(z)} & \textbf{MCE(z)} & \textbf{ECE(q)} & \textbf{MCE(q)} & \textbf{W-dist(z)} & \textbf{KLD(z)} & \textbf{W-dist(q)} & \textbf{KLD(q)} \\ \hline
\textbf{\begin{tabular}[c]{@{}c@{}}NLL Loss\\ (base)\end{tabular} \cite{loss_att} }     & 0.00224 & 0.00886 & 0.03503 & 0.12766 & 0.00225 & 0.00130 & 681.220   & 0.607 \\ \hline
\textbf{\begin{tabular}[c]{@{}c@{}}NLL Loss\\ (full)\end{tabular} \cite{loss_att}}     & 0.00146 & 0.02318 & 0.59936 & 0.77085 & 0.12145 & 0.07374 & 10953.997 & 1.768 \\ \hline
\textbf{\begin{tabular}[c]{@{}c@{}}Calibration\\  loss\end{tabular} \cite{bosch-calib} }   & 0.00163 & 0.01464 & 0.09681 & 0.30432 & 0.01529 & 0.01258 & 1501.167  & 0.848 \\ \hline
\textbf{\begin{tabular}[c]{@{}c@{}}fCal-WS\\ (ours)\end{tabular}}      & 0.00104 & 0.00656 & 0.00832 & 0.06299 & 0.00022 & 0.00015 & 97.245    & 0.201 \\ \hline
\textbf{\begin{tabular}[c]{@{}c@{}}fCal-KL\\ (ours)\end{tabular}}      & 0.00126 & 0.00880 & 0.01686 & 0.11125 & 0.00037 & 0.00025 & 304.647   & 0.403 \\ \hline
\textbf{\begin{tabular}[c]{@{}c@{}}Temperature\\  scaling\end{tabular} \cite{bosch-calib} } & 0.00226 & 0.00928 & 0.02705 & 0.10635 & 0.00206 & 0.00110 & 754.356   & 0.637 \\ \hline
\end{tabular} }
\vspace{0.3em}
\caption{Object detection - Cityscapes \cite{cordts2016cityscapes}: Results of \coolname{} and other baselines for various calibration metrics. }
\label{table:cs-calibration}
\end{table}
In tables \ref{table:kitti-consistency} and \ref{table:kitti-calibration}, we extensively report results for various metrics and baselines. We see that \coolname{} is able to obtain highly consistent and calibrated results. We observe that when we train entire model with NLL Loss\footnote{In practice, NLL Loss is trained without freezing any part of the model. In this work, we just train the uncertainty head with NLL loss to have base model, which is used as weight initializer for other methods. NLL Loss(base) works better than NLL Loss(full) for consistency and calibration. Hence in the main paper, we report results for the base model, as opposed to full model. But if we train entire model with loss attenuation as done in practice, due to its mean seeking nature, we observe that it yields extremely poor consistency.}(\cite{loss_att}), due to its mean seeking nature, it is trying to maximize the likelihood, and predicting very low uncertainty values, resulting in overconfident uncertainty estimates. This results in inconsistent predictions as evident from the mMAP, PDQ and PDQ(spatial) values. Note that many baselines have high consistency values but poorer calibration, which is a result of highly inflated uncertainty estimates. High consistency won't be very useful if we do not have good calibration. So consistency metrics must be interpreted in conjunction with calibration metrics, to make accurate conclusions. Currently, \coolname{} achieves state of the art calibration, while having competitive consistency. In table \ref{table:kitti-calibration}, we also report Maximum calibration error(MCE), Wasserstein distance and KLD, between proposed and target distributions.

\subsection{KITTI Depth Estimation}
\label{suppl:depth_results}

\begin{table}[tbh]
\begin{adjustbox}{width=\textwidth}
\begin{tabular}{c|c|c|c|c|c}
\hline
\textbf{Approach}                                & \textbf{SiLog}   & \textbf{RMSE}   & \textbf{ECE(z)} & \textbf{ECE(q)} & \textbf{NLL} \\ \hline
\textbf{NLL Loss\cite{loss_att}}                 & $9.213 \pm 0.092$  & $2.850 \pm 0.035$ & $2.39 \pm 0.224$  & $99.9 \pm 0.001$  & $3.403 \pm 0.258$  \\ 
\textbf{Calibration Loss\cite{bosch-calib}}      & $9.604 \pm 0.165$  & $2.918 \pm 0.015$ & $1.71 \pm 0.412$  & $99.9 \pm 0.000$  & $2.878 \pm 0.262$  \\ 
\textbf{Temperature Scaling\cite{bosch-calib}}   & $9.213 \pm 0.092$  & $2.850 \pm 0.035$ & $2.36 \pm 0.214$  & $99.9 \pm 0.004$  & $3.362 \pm 0.221$  \\ 
\textbf{\coolname{}-KL (ours)}                   & $9.679 \pm 0.091$  & $2.911 \pm 0.293$ & $0.074 \pm 0.021$ & $22.5 \pm 13.684$ & $2.004 \pm 0.143$  \\ 
\textbf{\coolname{}-Wass (ours)}                 & $9.509 \pm 0.098$  & $3.202 \pm 0.247$ & $0.156 \pm 0.044$ & $67.9 \pm 9.616$  & $2.157 \pm 0.159$  \\ \hline
\end{tabular}
\end{adjustbox}
\vspace{0.6em}
\caption{\textbf{Depth Regression - KITTI~\cite{kitti}}: 
\coolname{} on average gives better calibration performance in comparison with the baselines. ECE scores have been scaled by $100$ to enhance readability}
\label{table:kitti_depth_suppl}
\vspace{-0.2cm}
\end{table}

We evaluate \coolname{} and the baseline calibration methods using SiLog and RMSE for deterministic performance and ECE(z), ECE(q) and NLL for calibration performance. We run every experiment over 5 seeds to ensure reproducibility and establish statistical significance. Table. \ref{table:kitti_depth_suppl} shows the full results with both mean and standard deviation over the 5 seeds. From Table. \ref{table:kitti_depth_suppl}, we can see that \coolname{} models outperform the baseline calibration techniques on all the calibration metrics for similar deterministic scores. We also observe that unlike Bokeh and Object Detection, Temperature scaling struggles at calibrating uncertainties using a single scale/temperature parameter for depth estimation because of the complex nature of the task and the size of the dataset. Its also apparent that the values of ECE(q) are larger in comparison to the Bokeh and Object Detection results, this can be attributed to the non-i.i.d. nature of the nearby pixels in the depth estimation task which will make it difficult for the models to obtain a low ECE(q) score.

\vspace{0.5cm}

\section{\coolname{} code snippet}

\vspace{0.5cm}

This is a python implementation of \coolname{} for wasserstein distance as divergence minimization criteria. As we see, \coolname{} can be easily implemented using standard python packages such as PyTorch \cite{paszke2019pytorch} and NumPy \cite{harris2020array}.

\vspace{0.5cm}

\begin{lstlisting}[language=Python]
import torch
from numpy.random import default_rng
from . import model ## our model to be trained

dataset_name = 'my_dataset'
dataloader = torch.utils.data.DataLoader(dataset_name) ## MxK, MxN
inputs, gts = iter(dataloader) ## BxK, BxN

## training procedure
## forward pass
mu, std_dev = model(inputs)
gts, mu, std_dev = gts.flatten(), mu.flatten(), std_dev.flatten()

## residuals
residuals = (gts - mu) / std_dev
                                                                                
## constructing a chi-squared variable

rng = default_rng()
dof = 75	## degrees of freedom
cs_samples = 100 ## number of chi-squared samples                                                                                                
chi_sq_samples = []  ## this will have our chi-squared samples
for i in range(cs_samples):
	indices = rng.choice(len(mu), size = dof, replace = False)
	chi_sq_samples.append((residuals[indices]**2).sum*())
                                                                                
chi_sq_samples = torch.stack(chi_sq_samples)
mu1, mu2, var1, var2 = dof, chi_sq_samples.mean(), 2*dof, chi_sq_samples.var()
                                                                                
## loss computation
wasserstein_loss = ((mu1 - mu2)**2 + var1 + var2 - 2*(var1*var2)**0.5)
wasserstein_loss.backward()                                                                 
\end{lstlisting}

\newpage
\section{\coolname{} beyond Gaussians:}

\vspace{0.5cm}

In this work, we proposed a way to calibrate aleatoric uncertainty, which is modeled as Gaussian error. But in most real life settings, this assumption may not hold true. In such a case, we need to have methods to calibrate uncertainty for non-Gaussian setups too. In this section, we show that \coolname{} can be easily extended to non-Gaussian setups.

Given a mini-batch containing $N$ inputs $x_i$, a probabilistic regressor predicts $N$ sets of parameters $f_p(x_i) = \phi_i$ to the corresponding probability distribution $s(y_i; \phi_i)$. Define $h: \mc{Y} \times \Phi \mapsto \mc{Z}$ as the function that maps the target random variable $y_i$ to a random variable $z_i$, which follows a known canonical distribution. In case of $s$ being a Gaussian, $\phi_i \triangleq (\mu_i,\sigma_i)$, and we chose $h$ to be $\frac{y_i - \mu_i}{\sigma_i}$. Which results in residuals $\{z_1, z_2, \ldots,  z_N\}$ following a standard normal distribution. We choose our target distribution $Q$ to be a chi-squared distribution, and compute the empirical statistics of the residuals to fit a proposal distribution $P$ of the same family as $Q$. We define a variational loss function that minimizes the $f$-divergence between these two distributions. 

In case of non-Gaussian distribution, we need to define the transformation function $h$, such that our residuals $z_i$ follow a known canonical distribution. Here, we propose a way to construct $h$, through a series of transformations, such that final residuals are distributed as samples of a standard normal distribution. This will enable us to construct chi-squared hyper-constraints easily. Let $S(y; \phi) = \int_{-\infty}^y s(y', \phi)dy'$ be the Cumulative density function of the predicted probability distribution $s(y_i; \phi_i)$. Below, we revisit the definition of calibration, 

\begin{equation}
    p(Y \leq y | s(y)) = \int_{-\infty}^y s(y')dy' \text{  } \forall y \in \mc{Y} \nonumber
\end{equation}

\noindent
\begin{theorem}
\label{theorem1}
If $x$ is a univariate random variable with continuous and strictly increasing cumulative density function $F$, then transforming a random variable by its continuous density function always leads to the same distribution, the standard uniform\cite{embrechts2013note}. Hence, if $y = F(x)$ then y $\sim$ U[0,1].
\end{theorem}

Here, $p(Y \leq y | s(y))$ is cumulative density function $S(y; \phi)$. Hence, according to theorem \ref{theorem1} we know that $p(Y \leq y | s(y)) \sim U[0, 1]$. If we have a non-Gaussian modeling assumption, we transform the predictions to uniform distribution using theorem \ref{theorem1}, and then we take quantile function of a standard normal distribution($\mathcal{N}(0,1)$), to get standard normal residuals. After that, we follow same procedure to construct chi-squared hyperconstraints as presented in the main paper. To illustrate this, we provide a colab notebook with this supplementary, to illustrate applicability of theorem \ref{theorem1}. If $S(y; \phi_i)$ is cumulative density function of $s(y)$, and $F^{-1}_{\mathcal{N}}(.)$ is inverse CDF of a standard normal distribution, then our modified algorithm for non-Gaussian case can be expressed as below, 

\begin{algorithm}[h]
\SetCustomAlgoRuledWidth{\textwidth}  %
\SetAlgoLined
\SetKwInOut{KInput}{Input}
\DontPrintSemicolon
\KInput{Dataset $D$, probabilistic neural regressor, $f_p$, degrees of freedom $K$, batch size $N$, number of samples for hyper-constraint $H$}
\For{$i = 1 \ldots N$}{
$ \phi_i \gets f_p(x_i)$ \\
$z_i \gets F^{-1}_{\mathcal{N}}(S(y_i; \phi_i)) $ 
}
$C = \emptyset$ \tcp*{\textnormal{Samples from Chi-squared distribution}}
\For{$i = 1 \ldots H$}{
    \tcp{\textnormal{Create a chi-squared hyper-constraint}}
    $\displaystyle q_i \gets \sum_{j = 1}^{K} z_{ij}^2,  z_{ij} \thicksim \{z_1 \cdot \cdot \cdot z_N\}$ \\
    C.append($q_i$)
}
$P \gets \textnormal{Fit-Chi-Squared-Distribution}(C)$ \\
$\mathcal{L}_{\text{\coolname{}}} \gets D_f(P || \chi_K^2)$ \\
\Return $\mathcal{L}_{\text{\coolname{}}}$
\caption{\coolname{} for non-Gaussian uncertainties}
\label{algorithm-non-gaussian}
\end{algorithm}

Note that above algorithm requires the CDF function $S(.)$ to be continuous and differentiable. The rest of the pipeline is already differentiable and this can be implemented using standard autodifferentiation~\cite{paszke2019pytorch}. Below, we present a code-snippet for a case when uncertainties are modeled as Laplace distribution, 
\vspace{0.5cm}
\begin{lstlisting}[language=Python]
import torch
from numpy.random import default_rng
from . import model ## our model to be trained

dataset_name = 'my_dataset'
dataloader = torch.utils.data.DataLoader(dataset_name) ## MxK, MxN
inputs, gts = iter(dataloader) ## BxK, BxN

## training procedure
## forward pass
mu, b = model(inputs)
gts, mu, b = gts.flatten(), mu.flatten(), b.flatten()

## CDF of laplace distribution
uni_var = (gts <= mu) * 0.5 * torch.exp( (gts - mu) / b ) + \ 
		(gts > mu) * 0.5 * (1 - 0.5 * torch.exp(-(gts - mu) / b)

## this is for numerical stability
uniform_samples = torch.clamp(uni_var, 0.0000002, 0.9999998) 
                                                                                
## inverse CDF of standard normal distribution
residuals = 0.0 + 1.0 * torch.erfinv(2 * uniform_samples - 1) * np.sqrt(2)
                                                                                
## constructing a chi-squared variable
rng = default_rng()
dof = 75	## degrees of freedom
cs_samples = 100 ## number of chi-squared samples                                                                                
                                 
chi_sq_samples = []  ## this will have our chi-squared samples
for i in range(cs_samples):
	indices = rng.choice(len(mu), size = dof, replace = False)
	chi_sq_samples.append((residuals[indices]**2).sum*())
                                                                                
chi_sq_samples = torch.stack(chi_sq_samples)
mu1, mu2, var1, var2 = dof, chi_sq_samples.mean(), 2*dof, chi_sq_samples.var()
                                                                                
## loss computation
wasserstein_loss = ((mu1 - mu2)**2 + var1 + var2 - 2*(var1*var2)**0.5)
wasserstein_loss.backward()
\end{lstlisting}

\end{document}